\documentclass[runningheads]{llncs}

\usepackage{eccv}

\usepackage{eccvabbrv}

\usepackage{graphicx}
\usepackage{booktabs}

\usepackage[accsupp]{axessibility}  

\usepackage{hyperref}

\usepackage{orcidlink}

\usepackage{url}

\usepackage{graphicx}
\usepackage{amsmath}
\usepackage{wrapfig}
\usepackage{subcaption}
\usepackage[most]{tcolorbox}
\usepackage{multicol}
\usepackage{multirow}
\usepackage{float}
\usepackage{makecell}
\usepackage{booktabs}
\usepackage{xcolor}
\usepackage{soul}
\usepackage{colortbl}

\newcommand{\datasetname}{\textbf{Video2Reaction}}

\newif\ifmodificationready
\modificationreadytrue
\newcommand{\modification}[1]{\begingroup\ifmodificationready\else\color{blue}\fi #1\endgroup}

\title{Video2Reaction: Mapping Video to Audience Reaction Distribution in the Wild} 

\titlerunning{Video2Reaction}

\author{Trang Nguyen\inst{1}\orcidlink{0009-0001-0167-8775}\thanks{Corresponding author} \and
Sidong Zhang\inst{1}\orcidlink{0009-0007-1952-8488} \and
Shiv Shankar\inst{1} \orcidlink{https://orcid.org/0000-0003-1631-2570} \and
Gauri Jagatap\inst{2} \orcidlink{https://orcid.org/0000-0001-7499-2581} \and
Deepak Chandran\inst{2} \orcidlink{https://orcid.org/0009-0007-5306-9625} \and
Andrea Fanelli\inst{2} \orcidlink{https://orcid.org/0000-0001-9876-9050} \and
Madalina Fiterau\inst{1}}

\authorrunning{T.~Nguyen et al.}

\institute{University of Massachusetts Amherst, 130 Governors Drive, Amherst, MA, USA\\
\email{tramnguyen@umass.edu, sidongzhang@umass.edu, sshankar@umass.edu, mfiterau@cs.umass.edu} \and
Dolby Laboratories, 1275 Market Street, San Francisco, CA, USA\\
\email{Gauri.Jagatap@dolby.com, Deepak.Chandran@dolby.com, Andrea.Fanelli@dolby.com}}

\begin{document}

\maketitle

\begin{abstract}
Understanding and forecasting audience reactions to video content are crucial for improving content creation, recommendation systems, and media analysis. To enable audience reaction prediction and other content engagement applications, we introduce \textbf{Video2Reaction}, a multimodal dataset that maps short movie segments to a distribution of \textit{induced emotions} of viewers in the wild, as expressed through social media. \textbf{Video2Reaction} spans more than 10,000 videos and serves as a reliable benchmark as well as a training resource for audience reaction prediction. To enable cost-effective continuous annotations as reactions may change over time, we develop a two-stage multi-agent pipeline using only open-source LLMs, achieving 86\% correctness under blind human verification despite the inherently noisy and subjective nature of the task. We establish the first benchmark for video-to-reaction-distribution prediction in the wild and show that pretrained foundation video models fail in zero-shot settings, while finetuning transforms them into state-of-the-art predictors capable of modeling both full reaction distributions and dominant responses from video alone. However, the task remains challenging: even the strongest methods achieve only 77\% Top-3 F1 in dominant reaction prediction (LLaVA-Next), highlighting a substantial gap in modeling collective audience reaction. \modification{Dataset and code are available at our project page: \url{https://information-fusion-lab-umass.github.io/video2reaction-bench.github.io/}.}
\end{abstract}

\section{Introduction}



\begin{figure}[t]
    \centering    \includegraphics[width=\linewidth]{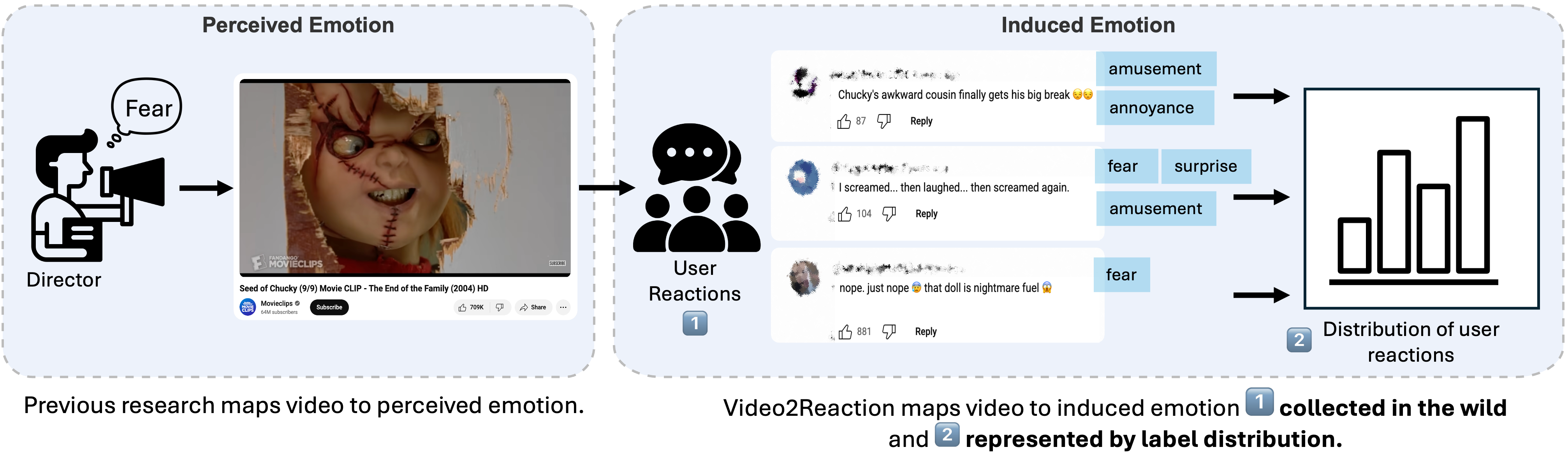}
    \caption{\textbf{Video2Reaction} is the first benchmark that uses video data to directly learn induced emotion distribution in the wild.}
    \label{fig:reaction_compare}
\end{figure}

Understanding how people react to video content is a crucial yet underexplored aspect of affective computing. Prior work has established a distinction between perceived emotion—the emotion conveyed or expressed by the content itself—and induced emotion—what is experienced by the audience \cite{tian2017recognizing}. While existing video sentiment datasets focus on perceived emotions (e.g., emotions of characters in the scene or filmmaker's emotion intent), audience reactions can vary greatly depending on personal, cultural, or temporal context. For example, Figure \ref{fig:reaction_compare} illustrates how a horror movie scene might be perceived as frightening or suspenseful based on the director's intent, but viewer reactions could diverge significantly-some may experience genuine fear and anxiety, others might laugh at predictable genre tropes, and those with personal trauma related to similar situations might be triggered and distressed. This underscores the need to study induced emotional reactions directly, rather than relying on perceived sentiment, so that the bias between the content creator's intention and possible audience feedback can guide the content creation. 

In this paper, we introduce \textbf{\datasetname}, a multimodal dataset consisting of over 10,000 videos of movie content, paired with audience reaction distribution derived from viewer comments, allowing for a precise mapping between visual content and the emotional reactions it induces. To our knowledge, \textbf{\datasetname} is the first and largest multimodal dataset capturing distribution of induced emotion from cinematic content in non-controlled environments (in the wild). Unlike perceived emotions, which are typically modeled as unimodal (single-label), induced emotions can vary from person to person, and it is important for content engagement applications to capture this variance rather than simply predict single or multi-label outputs. To address this, we frame audience emotion recognition as a \textbf{label distribution learning} (LDL) problem \cite{gengLabelDistributionLearning2016}, in which each video is mapped to a distribution of emotional reactions across viewers .


Our main contributions are as follows:
\begin{itemize}
\item \textbf{A large-scale benchmark for induced audience emotion in the wild.} To our knowledge, \textbf{\datasetname} is the first and largest multimodal dataset capturing \textit{induced emotion} in the wild (10,348 videos spanning approximately 400 hours and 800,000 comments) from cinematic content.
\item \textbf{A scalable, updatable LLM-based annotation pipeline with extensive quality analysis.} We develop a two-stage multi-agent LLM-based annotation pipeline that enables cost-effective, extensible reaction labeling, paving the way for future dataset updates as new content and evolving audience perspectives emerge. Beyond scalability, we conduct extensive quantitative and human evaluations to characterize annotation reliability, error modes, and robustness. Two complementary human evaluation results show LLM annotations operate within the range of human agreement while enabling large-scale distributional labeling at a fraction of the cost.
\item \textbf{A new task: distributional video-to-reaction prediction.} We propose a novel and challenging benchmark: predicting audience reaction distributions directly from multimodal video content, and design a comprehensive evaluation framework that captures both distributional prediction and dominant reaction classification. Our extensive benchmark, including classical LDL algorithms (i.e. SA-BFGS \cite{gengLabelDistributionLearning2016}), adapted multimodal emotion recognition models (i.e. CubeMLP \cite{sun2022cubemlp}), and large vision-language models (i.e. Gemini 2.5 \cite{gemini25report}), demonstrate that this task is challenging yet learnable.
\item \textbf{Demonstrating the potential of foundation models for audience reaction forecasting.} We demonstrate that pretrained VLMs fail zero-shot but can be transformed into state-of-the-art audience reaction predictors via lightweight finetuning, with preliminary evidence of cross-dataset transfer. 
\end{itemize}
\section{Related Work}
\subsection{Emotion Video Datasets}


\begin{table}[b]
\centering
\caption{Comparison of \textbf{Video2Reaction} and Existing Emotion Prediction Video Datasets}
\resizebox{\textwidth}{!}{
\begin{tabular}{lcccc}
\toprule
\textbf{Dataset} & 
\parbox{2.5cm}{\centering \textbf{\# Videos (Hours)}} & 
\parbox{2.5cm}{\centering \textbf{Emotion Type}} & 
\parbox{2.8cm}{\centering \textbf{Emotion Setting}} & 
\parbox{2.8cm}{\centering \textbf{Emotion Representation}} \\
\midrule
IEMOCAP \cite{busso2008iemocap}                  & 7,433 (12 hours)                     & Perceived                & Lab                               & Single-label                     \\ 
MELD \cite{poria2018meld}                     & 13,000 (13 hours)                     & Perceived              & Lab                               & Single-label                     \\ 
CMU-MOSEI \cite{zadeh2018multimodal}                & 23,453 (66 hours)                       & Perceived               & Lab                               & Continuous, Single-label         \\ 
LIRIS-ACCEDE \cite{baveye2015liris} \cite{muszynski2019recognizing}             & 9, 800 (27 hours)                        & Perceived, Induced                & Lab            & Continuous                       \\ 
COGNIMUSE \cite{zlatintsi2017cognimuse}                 & 50 (3.5 hours)                        & Induced                 & Lab              & Continuous, Single-label                       \\ 
DEAP \cite{koelstra2011deap}                     & 120 (2 hours)    & Induced                 & Lab             & Continuous                       \\ 
CMSV \cite{xu2024infer} & 8,210 (69 hours) & Induced & Social Media & Single-label \\
VCE \cite{mazeika2022would} & 61,046 (239 hours) & Induced & Lab & Distributional \\
\midrule
\rowcolor{gray!10}
\textbf{Video2Reaction} (ours) & 10,348 (398 hours) & Induced & Social Media & Distributional                     \\ 
\bottomrule
\end{tabular}
}
\label{tab:dataset_comparison}
\end{table}

Table \ref{tab:dataset_comparison} summarizes key characteristics of existing emotion video datasets and \textbf{\datasetname}. Most prior datasets like  IEMOCAP~\cite{busso2008iemocap}, MELD~\cite{poria2018meld}, and CMU-MOSEI~\cite{zadeh2018multimodal} focus on \textit{perceived emotions}—the emotions expressed or felt by the characters in a movie scene—rather than the \textit{induced emotions} experienced by the audience. Some datasets like LIRIS-ACCEDE \cite{baveye2015liris}, COGNIMUSE~\cite{zlatintsi2017cognimuse}, and DEAP~\cite{koelstra2011deap} have attempted to capture induced emotion. 
However, those datasets only capture emotional response in the 2D valence-arousal space, whereas our dataset captures the full distribution of categorical emotions. Furthermore, prior datasets often rely on controlled environments, such as lab settings or small groups of participants watching content together~\cite{muszynski2019recognizing, tian2017recognizing}, limiting the impact across larger and more diverse demographics. For instance, 
Muszynski \etal ~\cite{muszynski2019recognizing} extended \modification{LIRIS}-ACCEDE database \cite{baveye2015liris} to use aesthetic features from movie scenes to model induced emotions, but only with 10 participants in a co-viewing setting—unlike the typical media consumption nowadays. CMSV \cite{xu2024infer} is a recent benchmark that attempts to capture induced emotion in the wild but their task is to predict induced emotion given a pair of video and comments while our benchmark predict induced emotion from video content only. 

Additionally, existing emotion video datasets typically aggregate reactions into a single outcome label (e.g., a majority vote or mean score) for each clip~\cite{koelstra2011deap, baveye2015liris, zlatintsi2017cognimuse}, failing to reflect the diversity of viewer responses. In contrast, we model the full distribution of emotional reactions, using soft labels derived from real-world audience responses.

Closest to our work is the VCE dataset \cite{mazeika2022would}, which collects categorical emotional responses to video content. As noted by its authors, VCE relies on a relatively controlled crowdworker pool and does not aim to model reactions from a culturally diverse, population-scale audience. In contrast, our work captures naturally occurring, large-scale audience reactions by leveraging social media data, beginning with YouTube as a globally popular platform. This design better reflects the ecological diversity and inherently distributional nature of real-world audience engagement. Furthermore, VCE requires substantial manual annotation effort—reportedly involving 400 annotators—making large-scale expansion and iterative updates costly. Our approach instead employs an agentic annotation pipeline to streamline and automate labeling, enabling faster dataset updates and improved adaptability given the non-static nature of the task.

\subsection{Label Distribution Learning}


Label Distribution Learning (LDL) was formalized by \cite{gengLabelDistributionLearning2016} as a framework to address label ambiguity problems in tasks such as age estimation and sentiment prediction. Unlike traditional classification, which assumes a single ground truth label per instance, LDL represents each instance with a distribution over labels, capturing uncertainty and correlation among neighboring labels.


LDL methods are generally categorized into three families \cite{gengLabelDistributionLearning2016}: problem transformation (PT), algorithm adaptation (AA), and specialized algorithms (SA). Further details are included in Section \ref{sec:benchmark} and in Appendix \ref{apx:comparative_algorithms}. 

The benchmark we design builds on existing LDL methods, extending them to incorporate foundational Vision-Large Language Models as an additional class of algorithms capable of learning distribution alignment \cite{meister2411benchmarking}.


\setlength{\tabcolsep}{8pt}
\begin{table*}[t]
\centering
\caption{Finer-grained reaction categories (21) used in \textbf{\datasetname{}} grouped by sentiment.}
\begin{tabular}{lp{7.8cm}}
\toprule
\textbf{Sentiment Category} & \textbf{Finer-grained Reaction Categories} \\
\midrule
Positive & amusement, excitement, joy, caring, admiration, relief, approval \\
Negative & fear, nervousness, embarrassment, disappointment, sadness, grief,
disgust, anger, annoyance, disapproval \\
Ambiguous & realization, surprise, curiosity, confusion \\
\bottomrule
\end{tabular}
\label{tab:sentiment_mapping}
\end{table*}
\setlength{\tabcolsep}{6pt}

\section{\datasetname\ Dataset}
The \textbf{\datasetname} dataset has been developed to facilitate the prediction of audience reactions from short movie segments. Each instance consists of a movie clip paired with a probability distribution over possible audience reactions (listed in Table \ref{tab:sentiment_mapping}). To facilitate future research on raw video analysis, we provide YouTube video IDs for all clips. Additionally, we release preprocessed features, including state-of-the-art visual embeddings and audio embeddings extracted using pretrained models (more details in Appendix \ref{apx:data_preprocessing}).

\subsection{Data Collection} \label{sec:data_collection}
We curate movie clips from the CondensedMovies dataset~\cite{bain2020condensed} (licensed with CC BY 4.0), which contains licensed content from the \textit{Movieclips} YouTube channel. Each clip is a channel-curated movie segment capturing a key scene from the film. The use of licensed content improves the longevity of the dataset, as these clips are less likely to be removed from the platform. To ensure meaningful audience engagement, we retain only videos with a minimum of 10,000 views and at least 10 comments. The selected clips, originally uploaded between 2011 and 2019, are downloaded for further processing. Viewer comments on these videos extend through 2025, enabling the dataset to capture evolving audience perspectives and contemporary references, supporting a more comprehensive and generalizable benchmark for reaction prediction. \modification{Comments are collected and aggregated into a reaction distribution at the clip level; each clip is further decomposed into key frames for visual feature extraction (Appendix \ref{apx:data_preprocessing}), but both training and prediction remain at the clip level.}

\subsection{Comment-level Reaction Annotation} 
\label{sec:annotation_pipeline}
\begin{figure}
    \centering
    \includegraphics[width=\linewidth]{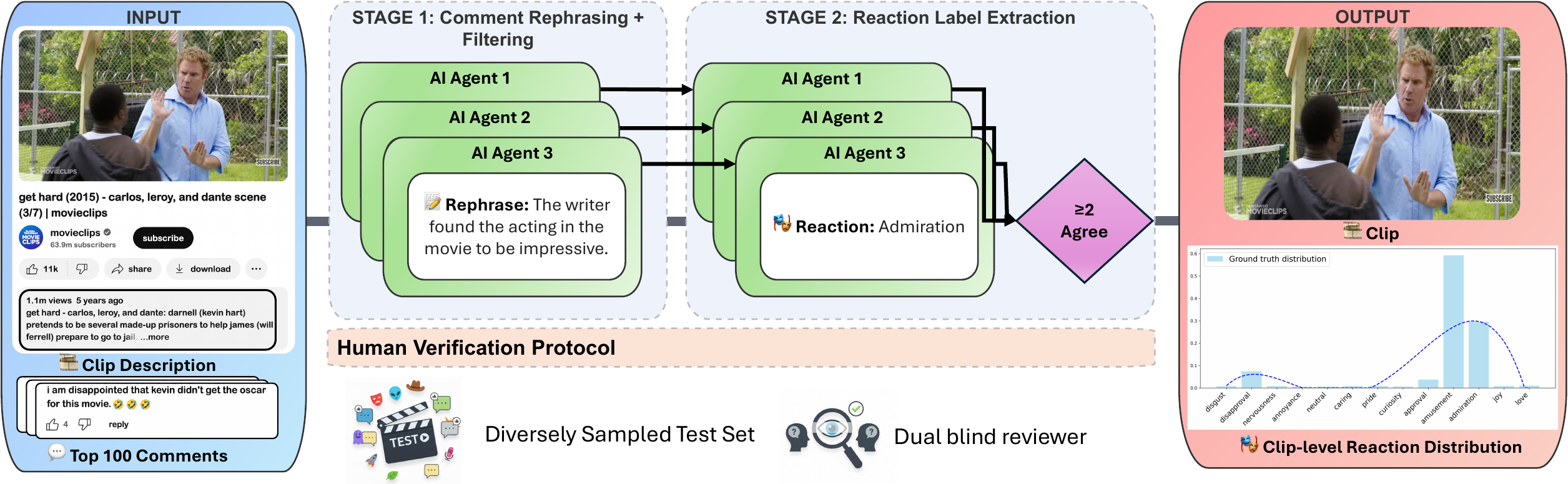}
    \caption{\textbf{Overview of \datasetname{} Two-Stage LLM-based Data Annotation Pipeline.} \\
    \underline{STAGE 1} rephrases comments to explicitly state their reactions towards the clip. It also filters out comments that lack a discernible reaction to the clip. \underline{STAGE 2} extracts reaction labels, with majority voting across three LLM agents to ensure consistency and discard ambiguous cases.
} 
    \label{fig:data_annotation_pipeline}
\end{figure}
\textbf{Reaction Taxonomy.}
To represent the complexity of audience reaction, we initially adopt the 28-category emotion taxonomy from GoEmotions \cite{demszky2020goemotions}, which was originally designed for Reddit comments and aligns naturally with our YouTube-based dataset. As a social media platform similar to Reddit, YouTube also features a wide range of audience interactions and emotional responses, making the GoEmotions taxonomy a natural fit. However, we drop 7 of the original categories in GoEmotions due to their significant under-representation in our data. These 7 reactions contribute to less than 0.01\% of the distribution mass on average. Our final taxonomy consists of 21 fine-grained emotions (Table \ref{tab:sentiment_mapping}). The definition for each category is in Appendix \ref{apx:reaction_definition}. 

\textbf{Two-stage Multi-agent Reaction Annotation Pipeline.} Given the volume of raw comments and the non-stationary nature of induced emotions over time, we designed a scalable and reliable multi-agent annotation pipeline that supports frequent dataset updates, ensuring long-term relevance and impact. Figure~\ref{fig:data_annotation_pipeline} outlines our LLM-based pipeline for annotating audience reactions based on user comments. Each comment is processed in two stages. The first stage of rephrasing comments is critical, as many audience comments reflect implicit reactions or off-topic remarks that need contextual interpretation to reveal their emotional intent. For example, a comment like \textit{“I’m so disappointed this actor didn’t win an Oscar”} will be mistakenly labeled as \textit{disappointment} without being rephrased as \textit{"the acting in the movie is so impressive"} in the first stage. The second stage then extracts relevant reaction labels from the rephrased comments from the first stage. Prompt details for both stages and qualitative comparison of single-stage versus two-stage annotations are provided in Appendix~\ref{apx:data_annotation}. \modification{Further validation on the Stage 1 filtering step shows that the pipeline achieves 0.83 accuracy with 1.00 sensitivity, confirming it does not discard relevant comments (details in Appendix~\ref{apx:data_annotation}).}

Motivated by prior findings in the use of multi-agent framework in text classification \cite{trad2024ensemble}, compound systems \cite{chen2024more}, and chain-of-thought reasoning \cite{wangself, choi2024multi}, we employ an ensemble of of three medium-sized multilingual instruction-tuned LLMs \footnote{LLaMA-3.1-8B-Instruct \cite{llama3herdmodels}
, Qwen2.5-14B-Instruct \cite{qwen2.5}
, and DeepSeek-R1-Distill-Qwen-7B \cite{deepseekai2025}} and adopt a straightforward majority voting approach for our emotion annotation task. This ensemble design enables inference-time speedup through agent parallelization, in contrast to annotator–critic architectures that require sequential interaction. Our choice to use three medium-sized LLMs with comparable performance aligns with prior observations that ensemble methods are most effective when constituent models exhibit similar strength \cite{trad2024ensemble}.

\textbf{Quality of Reaction Annotations.}
To assess the quality of annotations in \textbf{\datasetname}, we conduct two complementary evaluations. 

\textit{Human–LLM annotation alignment.} We perform independent human annotation of video–comment pairs with 29 participants and compute both inter-rater correlation and rater–LLM correlation following the protocol of GoEmotions \cite{demszky2020goemotions}. After filtering ambiguous cases, 233 comments remain with a median of 3 annotators per comments. We compute Spearman correlation per emotion between each rater and the mean of other raters, and between each rater and our LLM-based pipeline. Consistent with prior findings in GoEmotions \cite{demszky2020goemotions}, agreement is moderate and varies substantially across emotions. The mean inter-rater correlation across 21 emotions is 0.428 (std = 0.233), reflecting the subjective and fine-grained nature of induced emotion labeling. The LLM achieves a comparable mean rater-LLM correlation of 0.402 (std = 0.243), closely matching human–human agreement. These results indicate that LLM annotations operate within the same reliability range as human annotators, supporting the validity of our annotation pipeline despite the inherent noise of the task. Further error analysis and breakdown by emotion category can be found in Appendix \ref{sec:error_analysis}. \modification{Using the same evaluation protocol, an ablated single-stage pipeline achieves a lower rater–LLM correlation of 0.34, compared to 0.40 for our two-stage pipeline and 0.43 for inter-rater agreement, confirming that the rephrasing stage improves annotation reliability (full comparison in Appendix~\ref{apx:data_annotation}).}

\textit{Dual-blind human verification.} To further account for annotation noise from human raters in this task and to increase the test sample size without incurring significant increase in human evaluation time, we adopt a dual blind human verification protocol. First of all, we randomly sample 100 movie clips with balanced representation across all movie genres. From each clip, 10 comments are randomly selected, yielding a total of 1,000 comments for human evaluation. Then each comment is independently reviewed by two annotators. In cases of disagreement, a third annotator is consulted, and the final label is determined by majority vote. Overall, $86\%$ of the LLM-assigned reaction labels were judged to be correct, $7.8\%$ incorrect, and $6.2\%$ indeterminate (Table \ref{tab:reaction_annotation_quality} in Appendix \ref{sec:error_analysis}). Only $0.2\%$ randomly-labeled comments are judged to be correct, showing that human annotators do not exhibit confirmation bias. 

We further analyze LLM annotation errors for users' consideration and future improvements. As shown in ~\Cref{tab:error_by_genre} (Appendix~\ref{sec:error_analysis}), performance remains above 70\% across all genres. However, genres such as Comedy and Drama pose greater challenges due to their reliance on subtle cues (e.g., pop culture references), which can obscure emotional tone even for human annotators. \modification{Breaking down errors by emotion category instead (Table~\ref{tab:error_by_emotion} in Appendix~\ref{sec:error_analysis}), we find that \textit{embarrassment} and \textit{nervousness} are the most error-prone categories, likely due to their reliance on subtle tonal cues that are easily lost during rephrasing.}


\textbf{Discussion on Alternative Data Construction Approaches.}
An alternative approach to modeling audience reactions is to recruit participants to watch videos and report their emotional responses, as done in VCE dataset \cite{mazeika2022would}. While this design allows controlled measurement of reaction distributions, it departs from natural viewing conditions and typically relies on relatively small and demographically constrained participant pools. As acknowledged by the creators of the VCE dataset \cite{mazeika2022would}, their annotations are not intended to represent the full diversity of audience emotional responses and are primarily designed for studying whether deep networks can acquire cognitive empathy. Such datasets therefore do not capture large-scale, in-the-wild audience behavior. In addition, participant-based data collection is costly and inherently static, making it difficult to update as cultural context and audience reactions evolve over time.

Another alternative is to manually annotate social media comments. However, fine-grained emotion labeling is time- and cost-intensive (median 40.7 seconds per comment in our human evaluation study), rendering large-scale and continuously updated annotation impractical.

In contrast, leveraging naturally occurring social media reactions enables us to capture large-scale, ecologically valid audience responses. Our LLM-based annotation pipeline makes it feasible to maintain a non-static, updatable benchmark. We do not claim that social-media-derived reactions replace controlled lab studies; rather, they provide a complementary and population-level perspective that would be difficult to obtain through small-scale human annotation alone.



\subsection{Dataset Statistics} \label{sec:data_stats}
\begin{table}[h]
\centering
\caption{Descriptive statistics for video and comment data in \datasetname{}.}
\resizebox{\textwidth}{!}{%
\begin{tabular}{lccccc}
\toprule
\textbf{Category} & \textbf{Total} & \textbf{Min} & \textbf{Mean} & \textbf{Median} & \textbf{Max} \\
\midrule
\multicolumn{6}{l}{\textbf{Movie-level Statistics}} \\
\midrule
Inter-segment Chebyshev distance & 1,545 & 0.05 & 0.48 & 0.48 & 0.91 \\
in reaction distribution ([0.0,1.0]) & & & & & \\
\midrule
\multicolumn{6}{l}{\textbf{Clip-level Statistics}} \\
\midrule
Clip Duration (sec) & 389.81 hrs & 23.15 & 135.61 & 127.60 & 367.36 \\
Key Scenes & 455,226 & 16 & 43.99 & 39.00 & 176 \\
\midrule
\multicolumn{6}{l}{\textbf{Comment-level Statistics}} \\
\midrule
Raw Comments & 771,684 & 19 & 74.57 & 77.00 & 100 \\
Retained Comments & 252,462 & 10 & 24.40 & 22.00 & 100 \\
\bottomrule
\end{tabular}%
}
\label{tab:video_comment_stats}
\end{table}
Table~\ref{tab:video_comment_stats} summarizes \datasetname{} at the movie, clip, and comment levels. The dataset spans \modification{389} hours of video from 1,545 movies and includes 10,348 clips. On average, each clip is segmented into 44 scenes and is associated with 24 viewer comments used to construct the reaction distribution—substantially more than prior LDL benchmarks such as \textit{Twitter\_LDL} and \textit{Flickr\_LDL} ~\cite{yangLearningVisualSentiment2017}, which constructs a label distribution from 11 annotators only. 

Importantly, \datasetname{} captures substantial variation in audience reactions within the same movie, with a median Chebyshev distance of $0.48$ (out of $1.0$) between clip-level distributions of the same movie. This highlights the need for clip-level rather than movie-level modeling.  

\begin{figure}[ht]
    \centering

    \begin{subfigure}[t]{0.48\linewidth}
        \centering
        \includegraphics[width=\linewidth]{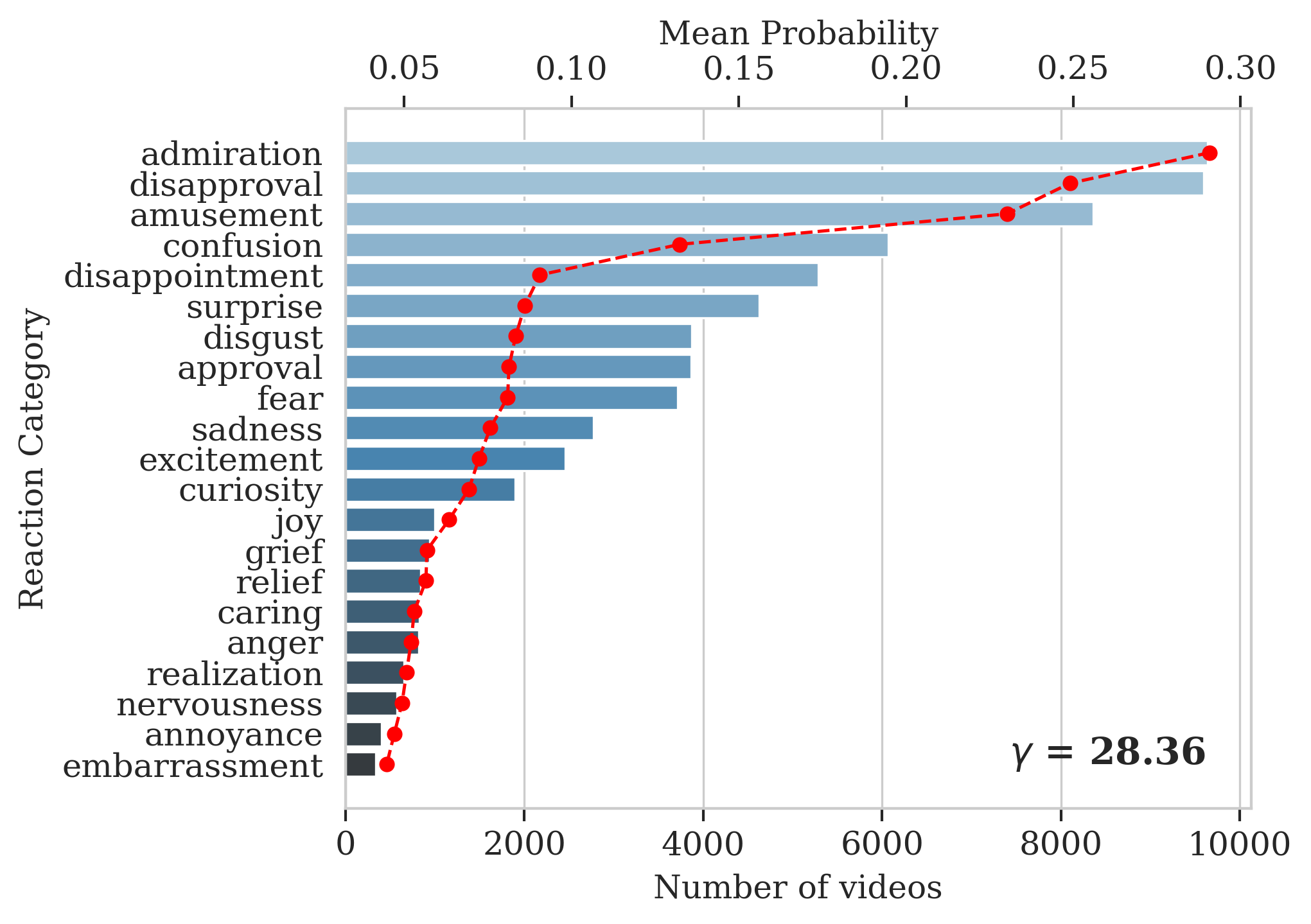}
        \caption{Total number of videos and mean video-level probability of each reaction category ($\gamma$ denotes imbalance factor)}
        \label{fig:reaction_distribution}
    \end{subfigure}
    \hfill
    \begin{subfigure}[t]{0.48\linewidth}
        \centering
        \includegraphics[width=\linewidth]{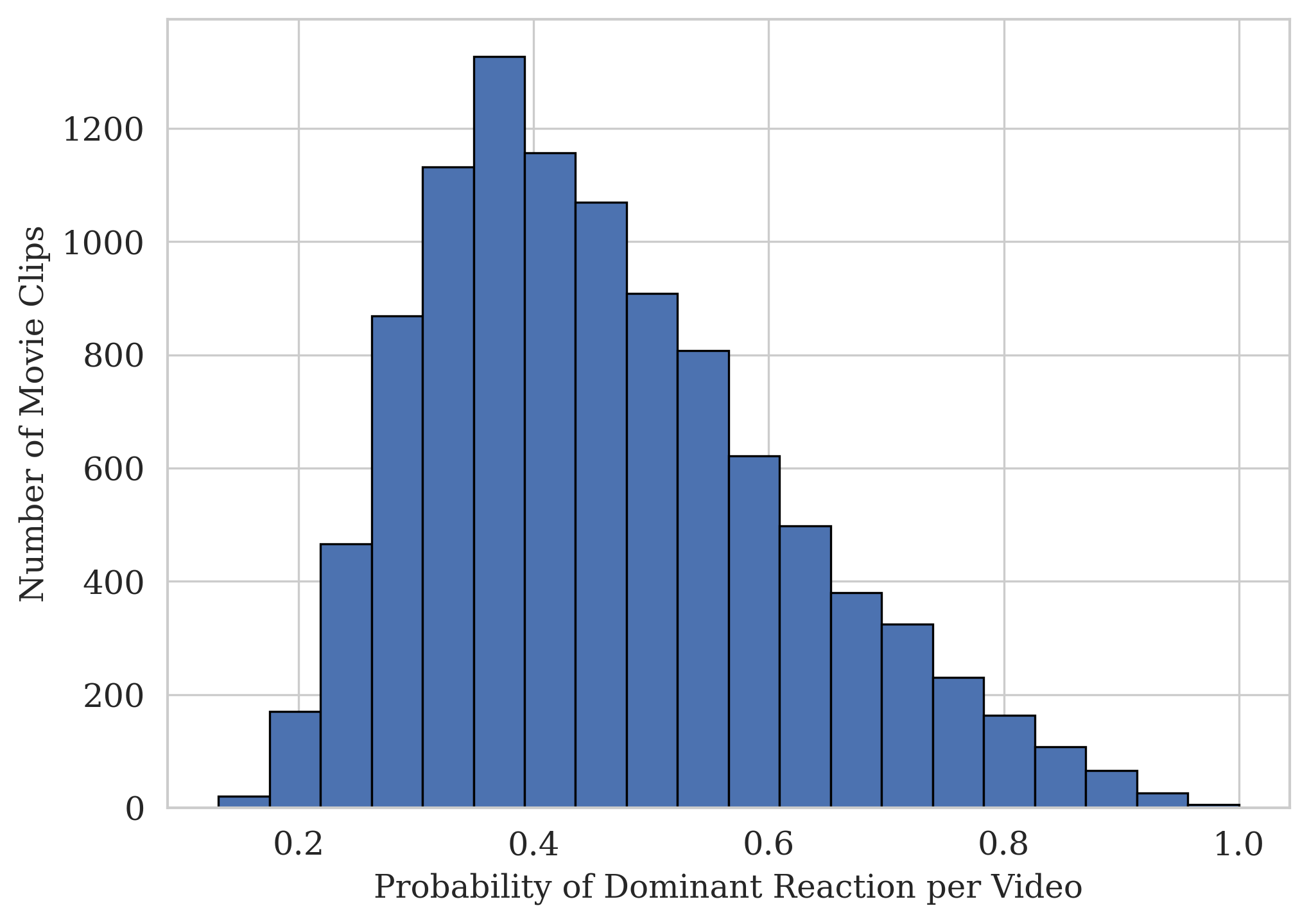}
        \caption{Distribution of dominant reaction probability per video}
        \label{fig:dominant_reaction_prob}
    \end{subfigure}
    \caption{Key Statistics on Reaction Outcome in the Video2Reaction dataset. }
    \label{fig:reaction_stats_combined}
\end{figure}

As shown in Figure~\ref{fig:reaction_distribution}, the dataset is highly imbalanced across the 21 reaction categories (with an imbalance factor of $28.36$), a challenge addressed in our evaluation design (Sections~\ref{sec:eval_metrics} and \ref{sec:exp_result}). Figure~\ref{fig:dominant_reaction_prob} further shows that top-1 reaction probability varies considerably across clips (with a median of approximately $0.4$), underscoring the importance of modeling full reaction distributions instead of predicting only the dominant label.


\begin{figure}[ht]
    \centering
    \includegraphics[width=\linewidth]{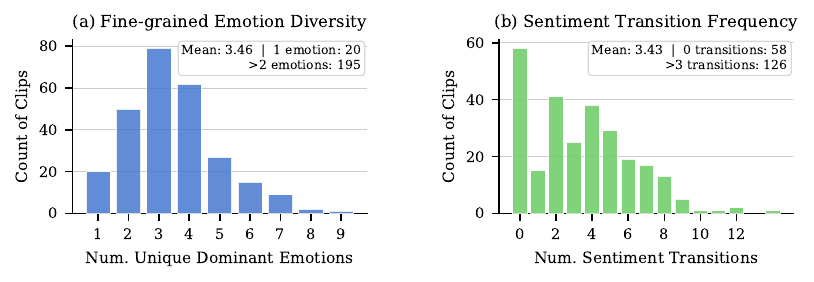}
    \caption{Distribution of fine-grained emotion and sentiment transitions across movie clips. Sentiment transitions are computed at the monthly level.}
    \label{fig:longti_setiment}
\end{figure}
\modification{
\subsection{Longitudinal Analysis of Audience Reactions} \label{sec:longitudinal}

Further analysis on how audience reactions to the same clip evolve over time shows that audience reactions are non-stationary and can shift meaningfully over time, motivating our scalable annotation pipeline. As shown in Figure~\ref{fig:longti_setiment}, clips exhibit an average of 3.46 distinct dominant emotions over time: only 7.5\% maintain a single dominant emotion, while 73.6\% exhibit more than three. At the sentiment level, clips show an average of 3.43 sentiment transitions, with 47.5\% exhibiting more than 3 transitions. For example, reactions to \textit{There Will Be Blood} shifted from predominantly positive in 2016 to disapproval by 2020.


}

\section{Audience Reaction Forecasting Benchmark} \label{sec:benchmark}

\subsection{Problem Formulation}
In this work, we frame the problem of audience reaction prediction as a label distribution learning (LDL) task. Unlike traditional classification, where the goal is to assign either a single label or a set of labels to each input instance, LDL seeks to predict a probability distribution over multiple labels, capturing the ambiguity and diversity in audience reactions.

Let $x$ denote an input video clip, which may contain visual, audio, and textual content. The audience reaction to $x$ is represented by a label distribution $\mathbf{d}_x = \{d_{xm}\}_{m=1}^M$, where $M$ is the number of affective reaction classes (e.g., amusement, confusion, fear, etc.), and each $d_{xm} \in [0, 1]$ indicates the proportion of annotators or viewers who associated label $m$ with the clip $x$. The label distribution satisfies the normalization constraint: $\sum_{m=1}^{M} d_{xm} = 1$. This distribution can be interpreted as the conditional probability $p(m|x)$ of observing reaction $m$ given video $x$. Our objective is to learn a model $f_\theta(x)$ that predicts a distribution $\hat{\mathbf{d}}_x$ approximating the empirical distribution $\mathbf{d}_x$.

Importantly, in our setting, $d_{xm}$ does not represent a soft target for a "correct" label in the traditional sense. Instead, it reflects the proportion of audience that has certain reaction given the video input. 

\vspace{-0.8em}
\subsection{Evaluation Metrics} \label{sec:eval_metrics}

Our benchmark is constructed along two complementary axes: (1) \textbf{full distribution evaluation}, which assesses how well the predicted distribution over all possible reactions matches the groundtruth distribution; and (2) \textbf{dominant reaction evaluation}, which focuses on how accurately the model identifies and estimates the probabilities of the most dominant reactions. The second axis is particularly relevant for real-world applications—such as content recommendation, moderation, or trailer editing—which depend primarily on anticipating the strongest or most likely viewer responses rather than capturing the full reaction spectrum. For each axis, we carefully select a suite of metrics designed to capture different types of errors that are important for our task, allowing future research to prioritize specific evaluation criteria based on downstream use cases. Table~\ref{tab:evaluation_metrics} summarizes all metrics used in the benchmark, along with their formulas and the error types they capture. 

\begin{table*}[t]
\centering
\caption{Summary of evaluation metrics used in the \datasetname{} benchmark. }
\resizebox{\textwidth}{!}{%
\begin{tabular}{p{1.5cm}ll p{7cm}}
\toprule
\textbf{Evaluation Category} & \textbf{Metric (Abbr.)} & \textbf{Formula} & \textbf{Error Type Captured} \\
\midrule
\multirow{6}{*}{\shortstack[l]{Full\\Distribution}}
& Chebyshev (Che) ↓ & $\max_j \left| d_j - \hat{d}_j \right|$ & Largest per-class prediction error (worst-case mismatch) \\
& Clark (Cla) ↓ & $\sqrt{ \sum_{j=1}^c \left( \frac{d_j - \hat{d}_j}{d_j + \hat{d}_j} \right)^2 }$ & Errors on rare reactions (small denominators amplified) \\
& Kullback-Leibler (KL) ↓ & $\sum_{j=1}^c d_j \ln \frac{d_j}{\hat{d}_j}$ & Underestimation of true reactions, false-zero sensitivity \\
& Cumulative Absolute Distance (CAD) ↓ & $\sum_j |CDF(d_j) - CDF(\hat{d}_j)|$ & Distribution shifts ignoring ordinal relations \\
& Cosine (Cos) ↑ & $\frac{ \sum_{j=1}^c d_j \hat{d}_j }{ \sqrt{\sum_{j=1}^c d_j^2} \sqrt{\sum_{j=1}^c \hat{d}_j^2} }$ & Directional mismatch of prediction vs. groundtruth \\
& Intersection (Inter) ↑ & $\sum_j \min(d_j, \hat{d}_j)$ & Non-overlapping reaction prediction error \\
\midrule
\multirow{3}{*}{\shortstack[l]{Dominant\\Reactions}}
& Mean Reciprocal Rank (MRR) ↑ & $\frac{1}{r}$ & Incorrect ranking of dominant reaction ($r$ is the predicted rank of the target dominant reaction) \\
& Top-1 Probability Error (TPE) ↓ & $\left| \hat{d}_{j^*} - d_{j^*} \right|, \quad j^* = \arg\max_j d_j$ & Probability misestimation of dominant reaction \\
& Top-$k$ F1 (weighted) ($F1_k$) ↑ & $\sum_{j} \frac{n_j}{N} \cdot \text{F1}_{j,k}$ & Both precision and recall-based errors for top-$k$ reactions\\
\bottomrule
\end{tabular}%
}
\label{tab:evaluation_metrics}
\end{table*}


\textbf{Full Distribution Evaluation.} Following \cite{gengLabelDistributionLearning2016}, we evaluate how closely the predicted reaction distribution matches the ground-truth distribution using a suite of statistical distribution distance metrics. Since some metrics (i.e. Canberra and Clark) capture very similar types of errors for our task, we decided to include in the benchmark three distance-based metrics—Chebyshev, Clark, Kullback-Leibler (KL) and two similarity-based metrics—Cosine similarity and Intersection. In addition, inspired by \cite{wen2023ordinal}, we incorporate an ordinal-aware evaluation by mapping the 21 reaction categories onto a valence-arousal-based emotional space. We then compute the Cumulative Absolute Distance (CAD) between the predicted and true \textit{ordered} distributions. The metric assigns lower penalties to misclassifications involving emotionally similar reactions and higher penalties to those involving emotionally distant categories, encouraging models to make semantically coherent predictions.

\textbf{Dominant Reaction Evaluation.} Unlike traditional emotion classification benchmarks, which often focus on unimodal distributions centered on a single perceived emotion, audience reactions in our task are multimodal, reflecting the diversity of viewer responses (as seen in Figure \ref{fig:dominant_reaction_prob}). Therefore, we evaluate not only the top-1 prediction (single-label) but also the task's performance as a multilabel classification problem. For single-label evaluation, we report the F1 score, Mean Reciprocal Rank (MRR), and Top-1 Probability Error (TPE). For multi-label evaluation, we compute F1 scores based on the top-k emotions from both the ground truth distribution and the model predictions. All F1 metrics are class-weighted to account for label imbalance.

\subsection{Comparative Models}

Building on the taxonomy of label distribution learning (LDL) algorithms proposed by~\cite{gengLabelDistributionLearning2016}, we organize comparative models into three categories—\textit{Problem Transformation}, \textit{Specialized Algorithms}, and \textit{Algorithm Adaptation}—and extend the framework by introducing a fourth category, \textit{Foundation VLMs}, to evaluate the potential of Video2Reaction as a finetuning dataset to improve the foundation models' capability of forecasting audience reaction distribution. 

\textbf{Problem Transformation (PT).}  
These methods reduce the LDL task to standard single-label learning (SLL) by resampling the training data. Each training instance $(\mathbf{x}_i, \mathbf{d}_i)$ with a label distribution $\mathbf{d}_i$ over $c$ classes is converted into multiple single-label examples $(\mathbf{x}_i, y_j)$ by sampling $y_j$ for class $j$ from $\mathbf{d}_i$. We include PT-Bayes~\cite{gengLabelDistributionLearning2016} and LDSVR~\cite{geng2015pre} as two representative baselines.

\textbf{Specialized Algorithms (SA).} 
These methods are explicitly designed for Label Distribution Learning (LDL), typically incorporating task-specific loss functions or optimization techniques. We evaluate three representative algorithms: SA-BFGS~\cite{gengLabelDistributionLearning2016}, which directly optimizes KL divergence using the BFGS algorithm; and LDL-LRR~\cite{ldl-lrr-jia2019label} and TLRLDL~\cite{kou122024exploiting}, both of which enhance the training objective by exploiting multi-label correlations. 

\textbf{Algorithm Adaptation (AA).}  
These models were originally designed for related tasks, such as multimodal sentiment analysis, and are adapted here for LDL. We include CubeMLP~\cite{sun2022cubemlp}, CTEN~\cite{zhang2023weakly}, and MMIM~\cite{han_improving_2021}. These models incorporate modules specifically designed to learn cross-modal features. 

\textbf{Foundation Vision-Language Models (VLMs).}  
Beyond the standard LDL taxonomy \cite{gengLabelDistributionLearning2016}, we introduce a fourth category to assess the zero-shot performance of large vision-language models. Although not trained for LDL, these models are increasingly used as proxies for human judgment in applications such as agent-based simulations~\cite{park2023generative} and behavioral research~\cite{hwang2023aligning, jiang2022communitylm}. We evaluate one commercial model, Gemini 2.5 Flash \cite{gemini25report}, and two leading open-source models—LLaVA-Next-Video-7B~\cite{zhang2024llavanextvideo} and Qwen2.5-VL~\cite{qwen2.5-VL}—in a prompted classification setting over fixed reaction labels.

\textbf{Implementation Details.} The PT, SA, and AA algorithms are trained on the dataset using 5 random seeds, and we report the mean of their performance in Tables~\ref{tab:full_distribution_eval} and~\ref{tab:dominant_reaction_eval} (standard deviation is reported in Appendix \ref{apx:full_results}. For foundation VLMs, we evaluate two variants. First, following \cite{meister2411benchmarking}, we apply temperature scaling on validation data to improve probability calibration. Second, we perform low-rank fine-tuning (LoRA) \cite{hu2022lora} with rank $r = 8$, $\alpha=16$, for 3 epochs. During fine-tuning, we incorporate taxonomy random reordering and synonym replacement to improve robustness to label-space variation. Further details on algorithm-specific implementation are listed in Appendix \ref{apx:comparative_algorithms}.

\section{Results} \label{sec:exp_result}


\subsection{Pretrained VLMs Fail to Predict Audience Reactions from Video Content}
\modification{Despite large-scale multimodal pretraining, VLMs fail to predict fine-grained audience reaction distributions in a zero-shot setting, even with temperature scaling. As shown in~\cref{tab:full_distribution_eval}, cosine similarity remains below 0.51 and intersection below 0.38, while Chebyshev distance exceeds 0.34 across models. Performance is similarly poor under dominant reaction evaluation (\cref{tab:dominant_reaction_eval}), with Top-1 F1 below 0.30 for all models. Thus, pretrained VLMs neither recover the overall reaction distribution nor reliably identify the dominant audience response. Although expected without task-specific supervision, the substantial performance gap despite large-scale video pretraining suggests that audience reaction forecasting is a challenging capability not learned through generic multimodal pretraining, motivating the need for dedicated datasets and benchmarks.}

\begin{table}[ht]
\centering
\caption{Full distribution benchmark results. Both finetuned foundation VLMs and SA-BFGS achieve best and comparable performances. Values in parentheses indicate the improvement of finetuning over the zero-shot temperature-scaled baseline. }
\resizebox{\textwidth}{!}{%
\begin{tabular}{lcccccc}
\toprule
\textbf{Model Name} & 
Cheb $\downarrow$ & 
KL $\downarrow$ & 
Cla $\downarrow$ & 
Cad $\downarrow$ & 
Inter $\uparrow$ & 
Cos $\uparrow$ \\
\midrule
\multicolumn{7}{l}{\textbf{Foundation VLMs}} \\
\midrule
Gemini 2.5 Flash & 0.3425 & 4.8102 & 3.4477 & 3.4316 & 0.3787 & 0.5029 \\
\multicolumn{7}{l}{LLaVA-Next-Video-7B} \\
- \textit{Temperature-scaled} & 0.4110 & 1.5547 & 3.9710 & 4.3269 & 0.2970 & 0.4185 \\
- \textit{LoRA Finetuned} & \textbf{0.1882} {\color{green!60!black} \scriptsize (-54.2\%)} & 3.1765 {\color{red} \scriptsize (+104.3\%)} & 3.9433 {\color{green!60!black} \scriptsize (-0.7\%)} & 2.1416 {\color{green!60!black} \scriptsize (-50.5\%)} & \textbf{0.6861} {\color{green!60!black} \scriptsize (+131.0\%)} & \textbf{0.8663} {\color{green!60!black} \scriptsize (+107.0\%)} \\
\multicolumn{7}{l}{Qwen2.5-VL} \\
- \textit{Temperature-scaled} & 0.3985 & 1.5216 & 3.9888 & 4.0333 & 0.3140 & 0.4401 \\
- \textit{LoRA Finetuned} & 0.2047 {\color{green!60!black} \scriptsize (-48.6\%)} & 3.4431 {\color{red} \scriptsize (+126.3\%)} & 3.9446 {\color{green!60!black} \scriptsize (-1.1\%)} & 2.2903 {\color{green!60!black} \scriptsize (-43.2\%)} & 0.6656 {\color{green!60!black} \scriptsize (+111.9\%)} & 0.8427 {\color{green!60!black} \scriptsize (+91.5\%)} \\
\midrule
\multicolumn{7}{l}{\textbf{Problem Transformation}} \\
\midrule
PT\_Bayes & 0.9668 & 21.1994 & 2.7268 & 6.9506 & 0.0144 & 0.0272 \\
LDSVR & 0.2584 & 4.9794 & \textbf{2.1272} & 2.8912 & 0.6146 & 0.7872 \\
\midrule
\multicolumn{7}{l}{\textbf{Specialized Algorithms}} \\
\midrule
SA\_BFGS & 0.2306 & \textbf{0.5976} & 3.9147 & 2.6711 & 0.6254 & 0.8089 \\
LDL\_LRR & 0.3293 & 2.2569 & 4.1227 & 3.5177 & 0.5242 & 0.7115 \\
TLRLDL & 0.3368 & 7.9606 & 3.2362 & 3.8317 & 0.4264 & 0.5968 \\
\midrule
\multicolumn{7}{l}{\textbf{Algorithm Adaptation}} \\
\midrule
CubeMLP & 0.2738 & 0.6900 & 3.9669 & 3.2122 & 0.5624 & 0.7513 \\
CTEN & 0.2432 & 0.6033 & 3.9542 & 2.8277 & 0.6071 & 0.7977 \\
MMIM & 0.2442 & 0.6076 & 3.9593 & 2.8548 & 0.6019 & 0.7946 \\
\bottomrule
\end{tabular}%
\label{tab:full_distribution_eval}
}

\end{table}
\subsection{Finetuning Transforms Foundation VLMs into State-of-the-Art Audience Reaction Predictors}
Under full distribution evaluation (\cref{tab:full_distribution_eval}), finetuned models reduce Chebyshev distance by roughly 50\% and nearly double cosine similarity. Intersection scores alos double, indicating substantially improved structural alignment with ground-truth distributions. Crucially, these gains extend to dominant reaction prediction. As shown in ~\cref{tab:dominant_reaction_eval}, finetuned VLMs models achieve MRR above 0.75 and Top-3 weighted F1 around 0.77. They consistently outperform all baselines across ranking and classification metrics. 

Video2Reaction also shows promise as a scalable pretraining resource for cross-dataset and cross-taxonomy transfer. Fine-tuning Qwen2.5-VL on a mixture of Video2Reaction and $1\%$ of VCE training data achieves $0.46$ Top-3 accuracy on the VCE benchmark \cite{mazeika2022would}, outperforming training on 10\% of in-domain data ($0.35$) (see Appendix \ref{apx:transfer_learning} for details). Although these gains remain far from fully supervised performance, they indicate meaningful cross-dataset transfer and suggest that Video2Reaction with automatic reaction annotation provides complementary supervision signals that can partially substitute expensive human-annotated data.

\begin{table*}[t]
\centering
\caption{Dominant Reaction Evaluation Benchmark Results. Best performance is shown in bold. Finetuned VLMs significantly outperforms all other baseline models.}
\resizebox{\textwidth}{!}{%
\begin{tabular}{lccccc}
\toprule
\textbf{Model Name} & TPE $\downarrow$ & MRR $\uparrow$ & F1 Top 1 $\uparrow$ & F1 Top 2 $\uparrow$ & F1 Top 3 $\uparrow$ \\
\midrule
\multicolumn{6}{l}{\textbf{Foundation VLMs}} \\
\midrule
Gemini 2.5 Flash & 0.3026 & 0.4378 & 0.2735 & 0.3332 & 0.3794 \\
\multicolumn{6}{l}{LLaVA-Next-Video-7B} \\
- \textit{Temperature-scaled} & 0.4103 & 0.1992 & 0.0143 & 0.1167 & 0.1374 \\
- \textit{LoRA Finetuned} & 0.1388 {\color{green!60!black} \scriptsize (-66.2\%)} & \textbf{0.7833} {\color{green!60!black} \scriptsize (+293.2\%)} & 0.6521 {\color{green!60!black} \scriptsize (+4460\%)} & \textbf{0.7374} {\color{green!60!black} \scriptsize (+531.8\%)} & 0.7672 {\color{green!60!black} \scriptsize (+458.3\%)} \\
\multicolumn{6}{l}{Qwen2.5-VL} \\
- \textit{Temperature-scaled} & 0.3923 & 0.3088 & 0.1958 & 0.2531 & 0.3037 \\
- \textit{LoRA Finetuned} & 0.1516 {\color{green!60!black} \scriptsize (-61.3\%)} & 0.7548 {\color{green!60!black} \scriptsize (+144.4\%)} & \textbf{0.6577} {\color{green!60!black} \scriptsize (+235.9\%)} & 0.7291 {\color{green!60!black} \scriptsize (+188.1\%)} & \textbf{0.7725} {\color{green!60!black} \scriptsize (+154.4\%)} \\
\midrule
\multicolumn{6}{l}{\textbf{Problem Transformation}} \\
\midrule
PT-Bayes & 0.9661 & 0.1535 & 0.0001 & 0.0031 & 0.0741 \\
LDSVR & 0.1599 & 0.7054 & 0.5034 & 0.5865 & 0.5696 \\
\midrule
\multicolumn{6}{l}{\textbf{Specialized Algorithms}} \\
\midrule
SA-BFGS & 0.1882 & 0.7163 & 0.5283 & 0.6075 & 0.6265 \\
LDL-LRR & 0.2496 & 0.6700 & 0.4965 & 0.5684 & 0.5803 \\
TLRLDL & 0.2759 & 0.5559 & 0.4516 & 0.4895 & 0.4858 \\
\midrule
\multicolumn{6}{l}{\textbf{Algorithm Adaptation}} \\
\midrule
CubeMLP & 0.2509 & 0.5996 & 0.2376 & 0.4399 & 0.5587 \\
CTEN & 0.2081 & 0.6939 & 0.4826 & 0.5950 & 0.5109 \\
MMIM & 0.2141 & 0.6749 & 0.4503 & 0.5728 & 0.5775 \\
\bottomrule
\end{tabular}%
}
\label{tab:dominant_reaction_eval}
\end{table*}



\subsection{Specialized LDL Methods Are Competitive on Distribution Metrics but Inferior on Dominant Reaction Prediction}
Classical LDL methods remain competitive under full distribution evaluation. Several specialized approaches achieve strong divergence-based scores, confirming their effectiveness at directly optimizing distributional objectives. However, this advantage does not translate to dominant reaction prediction. While methods such as SA-BFGS and LDSVR attain moderate MRR ($\sim$0.70), their Top-1 F1 scores consistently lag behind finetuned VLMs ($\sim$0.65 vs. $\sim$0.53). The gap persists across Top-2 and Top-3 metrics. However, LDL methods are substantially more computationally efficient than finetuned VLMs at both training and inference time. Thus, further research on these approaches remain attractive for resource-constrained settings.

Qualitative analysis (Appendix~\ref{apx:viz_samplewise_plot}) reveals a consistent pattern: specialized LDL models capture overall distribution shape but underestimate the leading reaction probability, particularly in low-entropy (highly unimodal) samples. 
\begin{table}[ht]
\centering
\caption{Reaction prediction performance across methods and input modalities.}
\label{tab:modality_comparison}
\resizebox{\textwidth}{!}{
\begin{tabular}{llcccc}
\toprule
\multirow{2}{*}{Method} & \multirow{2}{*}{Input Modality} &
\multicolumn{2}{c}{Full Distribution} &
\multicolumn{2}{c}{Dominant Reaction} \\
\cmidrule(lr){3-4} \cmidrule(lr){5-6}
 &  & Cheb $\downarrow$ & KL $\downarrow$ & MRR $\uparrow$ & F1$_1$ $\uparrow$ \\
\midrule

\multirow{3}{*}{SA-BFGS}
& Visual + Audio
& 0.314
& 0.907
& 0.552
& 0.306
\\

& Visual + Text
& 0.234
& \textbf{0.588}
& 0.709
& 0.515
\\

& Visual + Audio + Text
& \textbf{0.231}
& 0.598
& \textbf{0.716}
& \textbf{0.528}
\\
\midrule

\multirow{3}{*}{LLaVA-Next}
& Visual
& 0.218
& 4.148
& 0.731
& 0.586
\\

& Text
& 0.202
& 3.693
& 0.750
& 0.605
\\

& Visual + Text
& \textbf{0.188}
& \textbf{3.177}
& \textbf{0.783}
& \textbf{0.652}
\\

\bottomrule
\end{tabular}
}
\end{table}
\modification{
\subsection{Text Modality Provides Significant Performance Gain for both LDL and Foundation Models} \label{sec:multimodal_ablation}
To assess the contribution of each input modality, we conduct an ablation using two representative models, SA-BFGS and LLaVA-Next, trained on different combinations of visual, audio, and text (clip description) inputs. Table~\ref{tab:modality_comparison} shows that incorporating the text modality leads to substantial performance improvements for both models, while adding the audio modality to SA-BFGS provides only marginal gains over visual and text alone.
}

\section{Conclusion}
This paper introduces \textbf{Video2Reaction}, the first dataset and benchmark for predicting fine-grained induced audience reaction distributions towards movie content in the wild. This dataset and algorithms developed using it enables video creators to anticipate audience feedback prior to content release, potentially leading to improved content quality. Our benchmark is also supported by a scalable annotation pipeline that enables iterative updates to adapt to the non-stationary nature of audience engagement. We propose a comprehensive evaluation framework with two complementary axes, full distribution prediction and dominant reaction estimation, and conduct extensive experiments across four categories of algorithms. Our findings demonstrate that this task is challenging yet feasible: low-rank finetuning foundation VLMs can improve the base models' performance significantly on both distribution and classification benchmarks. 


\textbf{Limitations.} First, Video2Reaction derives reaction representations solely from YouTube, which may not capture broader audience behavior. Future work will extend the dataset to additional platforms (e.g., TikTok and Bilibili) using our scalable annotation pipeline. Demographic analysis is also not possible due to platform privacy policies. Second, like many fine-grained classification benchmarks, Video2Reaction exhibits a long-tail distribution that remains underexplored. Addressing rare reactions is an important direction for improving prediction performance. Finally, while preliminary results suggest cross-dataset and cross-taxonomy transfer (Appendix \ref{apx:transfer_learning}), a more systematic study is needed to understand how Video2Reaction can support foundation models that generalize across domains and taxonomies.



\subsubsection{Acknowledgements} The computational resources for this work are provided by the Unity Research Computing Platform, a multi-institutional cluster led by the University of Massachusetts and the University of Rhode Island. We also thank all the reviewers who provided quality review for our annotation pipeline.

\newpage
\bibliographystyle{splncs04}
\bibliography{manual_references, references}
\newpage
\appendix
\begin{center}
{\Large Video2Reaction: Mapping Video to Audience Reaction Distribution in the Wild \\
\vspace{1em}
Supplemental Material}
\end{center}

\section{Dataset Details}
\subsection{Dataset Metadata} \label{apx:metadata}
Table \ref{tab:dataset_metadata} summarizes key details of all features included in the dataset. Metadata features are recorded in the split-specific JSON files that can be downloaded at \url{https://huggingface.co/datasets/infofusionlab/Video2Reaction}. Preprocessed Features and Reaction Outcome can be loaded directly from Huggingface or by using our custom Python script.  

\begin{table}[ht]
\centering
\caption{Dataset Feature Details}
\begin{tabular}{lp{5cm}c}
\toprule
\textbf{Feature} & \textbf{Description} & \textbf{Shape} \\ 
\midrule
\multicolumn{3}{l}{\textbf{Metadata}} \\ 
\midrule
video\_id & YouTube video identifier & - \\
imdbid & IMDb movie identifier & -\\ 
genre & List of movie genres & - \\ 
country & List of movie countries & - \\ 
movie\_name & Name of the movie & - \\ 
clip\_name & Name of the clip & - \\ 
clip\_description & Description of the clip provided by the channel & - \\ 
\midrule
\multicolumn{3}{l}{\textbf{Preprocessed Features ($K$ denotes number of key frames)}} \\ 
\midrule
visual\_feature & ViT embeddings\footnote{\url{https://huggingface.co/google/vit-base-patch16-224}} of the middle frame of each scene & (K, 768) \\
audio\_acoustic\_feature & CLAP embeddings\footnote{\url{https://huggingface.co/laion/larger_clap_general/tree/main}}, pre-trained on a mixture of sounds & (K, 1024) \\
audio\_semantic\_feature & HuBERT embeddings\footnote{\url{https://huggingface.co/facebook/hubert-large-ll60k}}, pretrained on speech only data & (K, 1024) \\
clip\_description\_embedding & BERT-based text embeddings\footnote{\url{https://huggingface.co/google-bert/bert-base-uncased}} for clip description & (768,) \\ 
movie\_genre & One-hot encoding of movie genres & (23,) \\
\midrule
\multicolumn{3}{l}{\textbf{Reaction Outcome}} \\ 
\midrule
reaction\_distribution & Distribution of viewer reactions (21 categories) & (21,) \\ \bottomrule
\end{tabular}
\label{tab:dataset_metadata}
\end{table}

\subsection{Video Data Preprocessing} \label{apx:data_preprocessing}
Each movie clip in \textbf{\datasetname} is segmented into key scenes using PySceneDetect's content adaptive detection algorithm \footnote{\url{https://www.scenedetect.com/}}. For each scene, we extract the following features:
\begin{itemize}
    \item \textbf{Visual Features}: ViT embeddings \cite{vit-wu2020} \footnote{\url{https://huggingface.co/google/vit-base-patch16-224}}
of the middle frame of each scene.
\item \textbf{Audio Acoustic Features}: CLAP embeddings \cite{clap} \footnote{\url{https://huggingface.co/laion/larger_clap_general/tree/main}},
 pre-trained on a mixture of sounds.

\item \textbf{Audio Semantic Features}: HuBERT embeddings \cite{hubert}, \footnote{\url{https://huggingface.co/facebook/hubert-large-ll60k}}
pretrained on speech only data.
\end{itemize}

In addition to temporal audio-visual features, we also include \textbf{Clip Description}, a short clip description provided by the Youtube channel, preprocessed using BERT-based text embeddings \cite{bert-base-uncased}
. Detailed description of additional dataset metadata is available in Appendix \ref{apx:metadata}.

\subsection{Reaction Category Definition} \label{apx:reaction_definition}
Table \ref{tab:reaction_taxonomy} provides definitions for each reaction category included in Video2Reaction dataset. The definitions are from GoEmotions taxonomy~\cite{demszky2020goemotions}.

\begin{table}[h]
\centering
\caption{Video2Reaction Reaction Categories and Definitions. Definitions are copied from GoEmotions taxonomy \cite{demszky2020goemotions}}
\begin{tabular}{llp{5cm}}
\toprule
\textbf{Sentiment} & \textbf{Reaction Category} & \textbf{Definition} \\
\midrule
\multirow{7}{*}{Positive} & Amusement & Finding something funny or being entertained. \\ 
 & Excitement & Feeling of great enthusiasm and eagerness. \\ 
 & Joy & A feeling of pleasure and happiness. \\
 & Caring & Displaying kindness and concern for others. \\ 
 & Admiration & Finding something impressive or worthy of respect. \\ 
 & Relief & Reassurance and relaxation following release from anxiety or distress. \\ 
 & Approval & Having or expressing a favorable opinion. \\ 
 \midrule
\multirow{8}{*}{Negative} & Fear & Being afraid or worried. \\ 
 & Nervousness & Apprehension, worry, anxiety. \\ 
 & Embarrassment & Self-consciousness, shame, or awkwardness. \\ 
 & Disappointment & Sadness or displeasure caused by the nonfulfillment of one’s hopes or expectations. \\ 
 & Sadness & Emotional pain, sorrow. \\ 
 & Grief & Intense sorrow, especially caused by someone’s death. \\ 
 & Disgust & Revulsion or strong disapproval aroused by something unpleasant or offensive. \\ 
 & Anger & A strong feeling of displeasure or antagonism. \\
 & Annoyance & Mild anger, irritation. \\ 
 & Disapproval & Having or expressing an unfavorable opinion. \\ 
 \midrule
\multirow{4}{*}{Ambiguous} & Realization & Becoming aware of something. \\ 
 & Surprise & Feeling astonished, startled by something unexpected. \\ 
 & Curiosity & A strong desire to know or learn something. \\ 
 & Confusion & Lack of understanding, uncertainty. \\
 \bottomrule
\end{tabular}
\label{tab:reaction_taxonomy}
\end{table}

\subsection{Data Preprocessing Details}
Each movie clip in Video2Reaction is segmented into key scenes using PySceneDetect’s content
adaptive detection algorithm\footnote{\url{https://www.scenedetect.com/docs/latest/cli.html}}. The algorithm detects scene transition using rolling difference in HSL colorspace. We use a relatively low threshold $3.0$ to segment the scenes so a fast-paced plot scene like a jump scare will be split into multiple detected scenes due to changes in the HSL colorspace. Each scene is represented using the middle frame. 

\section{Data Annotation Pipeline}
\subsection{Implementation Details}
\label{apx:data_annotation}
Given the volume of raw comments to process, we employ an ensemble of three medium-sized multilingual instruction-tuned LLMs—LLaMA-3.1-8B-Instruct\footnote{\url{https://huggingface.co/meta-llama/Llama-3.1-8B-Instruct}}, Qwen2.5-14B-Instruct\footnote{\url{https://huggingface.co/Qwen/Qwen2.5-14B-Instruct}}, and DeepSeek-R1-Distill-Qwen-7B\footnote{\url{https://huggingface.co/deepseek-ai/DeepSeek-R1-Distill-Qwen-7B}}—chosen for their strong performance and favorable efficiency. All LLM agents share the same prompt for both stages, which are listed below.

The LLM annotation pipeline requires two following inputs:
\begin{itemize}
    \item Clip Description, to set context to understand the sentiment of the comments. Our pipeline uses the short clip description provided by @MOVIECLIPS Youtube channel but we can also use a description generated by a video understanding model if no existing description is available.
    \item Comment, user-written comments on youtube.
\end{itemize}

\clearpage
\begin{tcolorbox}[colback=gray!10!white, colframe=gray!30!black, title=Stage 1: Rephrase and Filter Comment Prompt]
You are to roleplay as a senior director speaking to a junior director in training. You are reviewing comments from audience members on a variety of scenes from a variety of movies/shows. You are explaining to the junior director what the audience member is likely feeling due to the clip along with your reasoning. The goal is to teach the junior director how film can predictably be used to invoke certain emotions; as a result, you should ignore comments from audience members where the analysis shows there is likely nothing the junior director can generalize.  
Be concise in explanations, limit them to 2 sentences at most.

\textbf{Example 1:}  
\texttt{<description>} Paul makes a pair of thieves pay for bringing a knife to a gun fight.  
\texttt{<comment>} TTC Subways, this is Toronto these days.  
\texttt{<explanation>} The audience member here is comparing how Toronto subways seems similar to the subways in the scene. You cannot generalize the feelings of this audience member broadly so we will ignore this comment.  
\texttt{<rephrased>} None  

\textbf{Example 2:}  
\texttt{<description>} Crocodile Dundee interrogates a gangster off the side of a building.  
\texttt{<comment>} It's Milton from Office Space.  
\texttt{<explanation>} The audience member just realizes the same actor from another TV show so there is no reaction towards any aspect of the clip here.  
\texttt{<rephrased>} None  

\textbf{Example 3:}  
\texttt{<description>} Marius mourns his fallen comrades.  
\texttt{<comment>} My favorite song of this masterpiece $\heartsuit$
\texttt{<explanation>} The audience member likes the music background in the movie clip so this is a reaction we want to know so that we can pay more attention to music and sound in the future.  
\texttt{<rephrased>} The writer loves the music background.  

\textbf{Example 4:}  
\texttt{<description>} The Thénardiers swindle guests at their inn.  
\texttt{<comment>} The 1985 version is the best one yet :)
\texttt{<explanation>} The audience member prefers another version of the movie but the comment does not explain why so we will ignore this comment.  
\texttt{<rephrased>} None  

\textbf{Example 5:}  
\texttt{<description>} Walking alone at night, Paul comes face-to-face with an armed criminal.  
\texttt{<comment>} Keep shooting til he's dead, leave no "victim" to identify you later..and sue you..  
\texttt{<explanation>} The audience member reiterates a character's action in the movie, implying that it is a good idea so this is an implicit reaction towards the character's decision or part of the plot in the scene.  
\texttt{<rephrased>} The writer agrees with the character's action in the scene.  

Now, analyze the following comment:  
\texttt{<description>} \{summary\}  
\texttt{<comment>} \{comment\}  
Director:
\end{tcolorbox}

\begin{tcolorbox}[colback=gray!10!white, colframe=gray!30!black, title=Stage 2: Extract Reaction Labels Prompt]
You are an assistant analyzing YouTube comments to extract audience reactions to a movie clip. Given the following inputs:  
- \textbf{clip\_description}: A short description of the movie clip.  
- \textbf{rephrased\_comment}: The original comment rewritten from a third-person perspective.  

Return a JSON object with the following fields:

\begin{itemize}
\item \texttt{high\_level\_reaction}: One or more words from \{"joy", "sadness", "anger", "surprise", "disgust", "fear", "neutral"\}.
\item \texttt{finer\_grained\_reaction}: One or more words from \{"admiration", "amusement", "anger", "annoyance", "approval", "caring", "confusion", "curiosity", "desire", "disappointment", "disapproval", "disgust", "embarrassment", "excitement", "fear", "gratitude", "grief", "joy", "love", "nervousness", "optimism", "pride", "realization", "relief", "remorse", "sadness", "surprise", "neutral"\}.
\item \texttt{reaction\_reason\_type}: One or more from \{"cinematography", "character", "acting", "sound and music", "editing and pacing", "narrative and thematic elements", "personal"\}; if no reason is clear, return \texttt{"none"}.
\end{itemize}

Return only a valid JSON object with these fields and \textbf{no additional text or explanations}.

\textbf{clip\_description}: \{clip\_description\} 

\textbf{rephrased\_comment}: \{rephrased\_comment\}  

\textbf{Output:}
\end{tcolorbox}

\subsection{Human Evaluation \& Additional Error Analysis}
\label{sec:error_analysis}

\textbf{Human-LLM Annotaiton Agreement.} Table \ref{tab:human_eval_emotion_corr} shows a comparison of inter-rater correlation and LLM-Human correlation by 21 emotion classes.

\begin{table}[t]
\centering
\caption{Comparison between human inter-rater agreement and LLM-human correlation across emotion categories.}
\begin{tabular}{lcc}
\hline
\textbf{Emotion} & \textbf{Inter-rater Corr.} & \textbf{LLM-Human Corr.} \\
\hline
admiration        & $0.622 \pm 0.158$  & $0.562 \pm 0.177$ \\
relief            & $0.792 \pm 0.063$  & $0.732 \pm 0.053$ \\
embarrassment     & $0.472 \pm 0.463$  & $0.719 \pm 0.380$ \\
curiosity         & $0.459 \pm 0.301$  & $0.690 \pm 0.281$ \\
confusion         & $0.643 \pm 0.235$  & $0.545 \pm 0.330$ \\
sadness           & $0.847 \pm 0.235$  & $0.524 \pm 0.244$ \\
anger             & $0.511 \pm 0.238$  & $0.484 \pm 0.399$ \\
disapproval       & $0.467 \pm 0.243$  & $0.445 \pm 0.245$ \\
surprise          & $0.279 \pm 0.349$  & $0.439 \pm 0.383$ \\
grief             & $0.740 \pm 0.307$  & $0.437 \pm 0.277$ \\
realization       & $0.292 \pm 0.289$  & $0.381 \pm 0.389$ \\
amusement         & $0.431 \pm 0.288$  & $0.380 \pm 0.290$ \\
joy               & $0.331 \pm 0.253$  & $0.285 \pm 0.232$ \\
excitement        & $0.162 \pm 0.270$  & $0.275 \pm 0.421$ \\
disgust           & $0.434 \pm 0.339$  & $0.265 \pm 0.330$ \\
disappointment    & $0.186 \pm 0.266$  & $0.191 \pm 0.404$ \\
annoyance         & $0.368 \pm 0.289$  & $0.165 \pm 0.236$ \\
approval          & $0.495 \pm 0.126$  & $0.070 \pm 0.176$ \\
fear              & $-0.043 \pm 0.013$ & $0.903 \pm 0.097$ \\
caring            & $-0.042 \pm 0.013$ & $-0.022 \pm 0.000$ \\
nervousness       & $0.541 \pm 0.171$  & $-0.031 \pm 0.006$ \\
\hline
\textbf{Mean} & $\mathbf{0.428}$ & $\mathbf{0.402}$ \\
\hline
\end{tabular}

\label{tab:human_eval_emotion_corr}
\end{table}
\noindent
\textbf{Dual-blind Human Verification.} We provide a summary of human evaluation on our test set in Table \ref{tab:reaction_annotation_quality}. We further analyze different types of errors that our annotation pipeline tend to make and present them in Table \ref{tab:error_by_emotion} and \ref{tab:error_by_genre}.


\begin{table}[ht]
\centering
\caption{Human evaluation of automated reaction annotation on a sample of 1000 comments.}
\begin{tabular}{lp{6cm}l}
\toprule
\textbf{Human Rating} & \textbf{Description} & \textbf{\# comments (\%)} \\
\midrule
Correct     & Most annotators agree the LLM-assigned labels are correct.      & 860 (86.0\%) \\
Incorrect   & Most annotators agree the LLM-assigned labels are incorrect.    & 78 (7.8\%)  \\
Not Sure    & Most annotators are unsure about the correctness of the labels (i.e. due to lack of context on movie references). & 62 (6.2\%)  \\
\bottomrule
\end{tabular}
\label{tab:reaction_annotation_quality}
\end{table}

\begin{table}[h]
\centering
\caption{Annotation performance across different movie genres.}
\begin{tabular}{lrrr}
\hline
\textbf{Movie Genre} & \textbf{\% Correct} & \textbf{\% Not Sure} & \textbf{\% Incorrect} \\
\hline
Adventure   & 96.23 &  1.89 &  1.89 \\
Fantasy     & 96.23 &  1.89 &  1.89 \\
Film-Noir   & 95.00 &  5.00 &  0.00 \\
Musical     & 93.94 &  0.00 &  6.06 \\
Documentary & 90.48 &  4.76 &  4.76 \\
History     & 90.20 &  3.92 &  5.88 \\
Sci-Fi      & 89.58 &  4.17 &  6.25 \\
Biography   & 88.24 &  5.88 &  5.88 \\
Romance     & 87.27 &  3.64 &  9.09 \\
Crime       & 87.04 &  7.41 &  5.56 \\
Horror      & 86.79 &  3.77 &  9.43 \\
Sport       & 84.09 &  4.55 & 11.36 \\
Thriller    & 83.33 &  1.85 & 14.81 \\
Mystery     & 83.02 &  3.77 & 13.21 \\
Music       & 81.63 & 16.33 &  2.04 \\
Animation   & 80.77 & 11.54 &  7.69 \\
War         & 80.39 &  5.88 & 13.73 \\
Family      & 80.00 &  7.27 & 12.73 \\
Comedy      & 79.25 & 16.98 &  3.77 \\
Drama       & 77.78 & 14.81 &  7.41 \\
Action      & 72.55 & 15.69 & 11.76 \\
\hline
\end{tabular}

\label{tab:error_by_genre}
\end{table}

\begin{table}[h]
\centering
\caption{Annotation performance across different emotion categories.}
\begin{tabular}{lrrr}
\hline
\textbf{Emotion Category} & \textbf{\% Correct} & \textbf{\% Not Sure} & \textbf{\% Incorrect} \\
\hline
annoyance      & 100.00 &  0.00 &  0.00 \\
relief         & 100.00 &  0.00 &  0.00 \\
confusion      &  94.64 &  5.36 &  0.00 \\
approval       &  93.94 &  0.00 &  6.06 \\
curiosity      &  92.31 &  0.00 &  7.69 \\
amusement      &  91.53 &  5.29 &  3.17 \\
admiration     &  90.41 &  5.17 &  4.43 \\
surprise       &  85.71 &  2.86 & 11.43 \\
disapproval    &  84.71 &  7.85 &  7.44 \\
excitement     &  82.35 & 17.65 &  0.00 \\
disappointment &  73.91 &  8.70 & 17.39 \\
disgust        &  72.73 &  9.09 & 18.18 \\
sadness        &  66.67 & 20.00 & 13.33 \\
fear           &  66.67 & 11.67 & 21.67 \\
realization    &  66.67 &  0.00 & 33.33 \\
joy            &  50.00 & 16.67 & 33.33 \\
caring         &  50.00 & 25.00 & 25.00 \\
grief          &  50.00 & 16.67 & 33.33 \\
anger          &   0.00 & 75.00 & 25.00 \\
embarrassment  &   0.00 & 33.33 & 66.67 \\
nervousness    &   0.00 &  0.00 &100.00 \\
\hline
\end{tabular}
\label{tab:error_by_emotion}
\end{table}

\modification{
\noindent
\textbf{Validation of Stage 1 Filtering.} To assess whether Stage 1 discards comments inappropriately, we randomly sample 100 comments (50 filtered out and 50 retained by the pipeline) and obtain blind human judgments on whether each comment should have been filtered. The pipeline achieves an accuracy of $0.83$ with $1.00$ sensitivity, indicating that it successfully avoids discarding comments with a discernible reaction. Specificity is lower ($0.77$), reflecting a more aggressive filtering strategy; this tradeoff is preferable for large-scale annotation, where minimizing noisy or irrelevant comments matters more than maximizing comment retention.

\noindent
\textbf{Single-stage vs. Two-stage Annotation Pipeline.} To quantify the benefit of the rephrasing stage, we compare our two-stage pipeline against an ablated single-stage variant that extracts reaction labels directly from the raw comment, skipping Stage 1. Using the same evaluation protocol as the human--LLM annotation alignment experiment (Section~\ref{sec:annotation_pipeline}), the single-stage pipeline achieves a mean rater--LLM correlation of $0.34$, compared to $0.40$ for the two-stage pipeline and $0.43$ for inter-rater agreement. This indicates that rephrasing comments to make their reaction explicit before label extraction substantially improves alignment with human judgments, validating our two-stage design.
}
\clearpage

\section{Implementation details on Comparative Algorithms} \label{apx:comparative_algorithms}

\subsection{Problem Transformation and Specialized Algorithms} \label{apx:pt_sa}
All PT and SA algorithms are implemented using PyLDL \footnote{\url{https://github.com/SpriteMisaka/PyLDL/tree/main}} Python Package. Since these methods are not designed to leverage temporal features, we apply average pooling to aggregate audio and visual inputs before feeding them into the models. For LDL-LRR \cite{ldl-lrr-jia2019label} and TLRLDL \cite{kou122024exploiting}, as recommended, we apply StandardScaler feature preprocessing and perform hyperparameter tuning on weight of different loss and regularization terms on the validation set.

\subsection{Adapted Algorithms} \label{apx:aa}




\begin{table*}[ht]
\centering
\caption{Model-specific Hyperparameters for Algorithm Adaptation (AA)}
\begin{tabular}{cc}
\toprule
\textbf{Model} & \textbf{Hyperparameter}  \\
\midrule
CubeMLP & \makecell{
encoders=lstm\\
d\_common=128\\
activate=gelu\\
time\_len=16\\
d\_hiddens=[[16, 2, 128],[8, 2, 128],[4, 1, 128]]\\
d\_outs = [[2, 2, 128],[2, 2, 128],[2, 2, 2]]\\
dropout=0.1\\
features\_compose\_t="cat"\\
features\_compose\_k="cat"
}\\
\midrule
CTEN & \makecell{n\_classes=21 \\ seq\_len=16 \\ audio\_embed\_size=1024 \\ visual\_embed\_size=768} \\
\hline
MMIM & \makecell{alpha=0.1 \\ beta=0.1 \\ update\_batch=1 \\ clip=1.0 \\ dropout\_a=0.1 \\ dropout\_v=0.1 \\ dropout\_prj=0.1 \\ n\_layer=1 \\ cpc\_layers=1 \\ d\_vout=16 \\ d\_aout=16 \\ d\_tout=16 \\ d\_tfeatdim=768 \\ d\_afeatdim=1024 \\ d\_vfeatdim=768 \\ n\_class=21 \\ d\_prjh=128 \\ pretrain\_emb=768 \\ mmilb\_mid\_activation=relu \\ mmilb\_last\_activation=tanh \\ cpc\_activation=tanh} \\
\hline
\end{tabular} \label{tab:model_hp}
\end{table*}

\begin{table*}[ht]
\centering
\caption{Adapted Algorithms Training Hyperparameters}
\begin{tabular}{|c|c|}
\hline
\textbf{Hyperparam} & \textbf{Value}  \\
\hline
Learning rate & 1e-3 \\
Weight decay & 1e-4 \\
Momentum & 0.9 \\
\#Epochs & 200 \\
Batch size & 128 \\
StepLR step size & 50 \\
StepLR gamma & 0.5 \\
\hline
\end{tabular} \label{tab:train_hp}
\end{table*}

\textbf{CubeMLP} \cite{sun2022cubemlp} combines temporally aligned unimodal features and mixes them across time, feature, and modality dimension using a structure consisting of 3 independent MLP units. 

\noindent\textbf{CTEN} \cite{zhang2023weakly} incorporates both uni-modal and cross-modal temporal attention mechanisms over visual and audio features extracted from key snippets, and further introduces an erasing strategy to localize context- and audio-relevant information in a weakly supervised setting. The major modification is that we only forward the processed visual and audio features to the model, so the ResNet encoders in the original CTEN to process raw modalities are removed. Further, since the erasing module requires access to the full original video, which is unavailable for \datasetname, we report benchmark results using CTEN without the erasing component. For the applicable hyperparamters, we follow the default values proposed in \cite{zhang2023weakly} to construct the model, and report them in Table \ref{tab:model_hp}.

\noindent\textbf{MMIM} \cite{han_improving_2021} introduces a two-stage end-to-end pipeline that learns fused representations from refined cross-modal features. The model is trained jointly to optimize both downstream task performance and mutual information between the fused representation and the unimodal input. In the first stage, mutual information is estimated by modeling a Gaussian mixture over positive and negative group. To assign samples to either group on \datasetname, we compute the summed probabilities of all positive and negative emotions for each sample, assigning it to the group with the dominant emotion category. Since the inputs are processed latent features, we replace the RNN models in the original MMIM with simple fully connected layers to align the latent dimension. For the applicable hyperparamters, we follow the default values proposed in \cite{han_improving_2021} to construct the model, and report them in Table \ref{tab:model_hp}.

For the three models, we use Pytorch with 1 GPU and use the same schedule to train them for better performance comparison. We use cross entropy loss between the output logits and the ground truth label distributions as the training loss, use SGD as the optimizer, and use StepLR as the optimizer scheduler. Details on the hyperparameters are reported in Table \ref{tab:train_hp}. Additionally, when training MMIM, it requires 2 separate training stages, and we use the same set of hyperparameters in Table \ref{tab:train_hp} for the two optimizers and schedulers in the two stages. Also, when training MMIM, it uses a single likelihood maximization loss in the first stage, and requires extra Contrastive Predictive Coding score and Gaussian mixture based mutual information estimation in the loss, besides the regular cross entropy loss we apply to every model.

\subsection{Foundation Vision-Large Language Models} \label{apx:vlm}

To extract the model's estimated probability of each reaction, we provide the VLMs a prompt with video input, a short description of the video, and a taxonomy of reaction categories. We then use the next token's probability as the predicted probability of each reaction category. The prompt is provided below:

\begin{tcolorbox}[colback=gray!10!white, colframe=gray!30!black, width=\textwidth, boxrule=0.5pt, arc=3pt, title={Prompt for Audience Reaction Forecasting using Foundation VLMs}, fonttitle=\bfseries, breakable]

\textbf{System Instruction}

You are an audience member watching a movie clip. You will be provided with the video, 
some context about the scene in the video, and a taxonomy of possible emotional reactions.

\textbf{Video Input}

[Movie clip]

\textbf{Scene Context}

In this scene, \textit{<clip\_description>}.

\textbf{Emotion Taxonomy}

A. curiosity \\
B. admiration \\
C. sadness \\
D. embarrassment \\
E. grief \\
F. realization \\
G. approval \\
H. caring \\
I. disgust \\
J. relief \\
K. confusion \\
L. nervousness \\
M. annoyance \\
N. joy \\
O. amusement \\
P. fear \\
Q. anger \\
R. disapproval \\
S. excitement \\
T. disappointment \\
U. surprise

\textbf{Question}

After watching the movie clip, how would you feel?  
Choose \textbf{one letter} corresponding to the emotion that best describes your reaction.

\textbf{Answer Format}

Answer: <letter>

\end{tcolorbox}

We use Low-Rank Adaptation (LoRA) \cite{hu2022lora} to finetune Qwen2.5-VL and LLaVA-Next with rank $r=8$ and scaling factor $\alpha=6$. 
Models are finetuned for 3 epochs using 1 A100 (80GB VRAM) and a batch size of 2 with gradient accumulation of 4 (effective batch size 8), a learning rate of $2\times10^{-4}$ with 100 warmup steps, and cross-entropy loss; we additionally apply text augmentation through synonym replacement ($p=0.3$) and taxonomy option reordering ($p=0.5$) during training.

\section{Additional Benchmark Results}
\subsection{Full results with standard deviation}
Table \ref{tab:full_distribution_eval_apx} and \ref{tab:dominant_reaction_eval_apx} are the same results as Table \ref{tab:full_distribution_eval} and \ref{tab:dominant_reaction_eval} in the main manuscript but with standard deviation across 5 runs. For VLMs, since it is expensive to finetune multiple times and the learning curve looks stable, we only record the performance after 1 run of finetuning.

\label{apx:full_results}
\begin{table*}[htb!]
\centering
\caption{Full Distribution Evaluation Benchmark Results.}
\resizebox{\textwidth}{!}{%
\begin{tabular}{lcccccc}
\toprule
\textbf{Model Name} & 
Cheb $\downarrow$ & 
KL $\downarrow$ & 
Cla $\downarrow$ & 
Cad $\downarrow$ & 
Inter $\uparrow$ & 
Cos $\uparrow$ \\
\midrule
\multicolumn{7}{l}{\textbf{Foundation Video Models}} \\
\midrule
Gemini 2.5 Flash (temperature-scaled) & 0.3425 & 4.8102 & 3.4477 & 3.4316 & 0.3787 & 0.5029 \\
LLava-Next-Video-7B (temperature-scaled) & 0.4110 & 1.5547 & 3.9710 & 4.3269 & 0.2970 & 0.4185 \\
Qwen2.5-VL (temperature-scaled) & 0.3985 & 1.5216 & 3.9888 & 4.0333 & 0.3140 & 0.4401 \\
Qwen2.5-VL (finetuned) & \textbf{0.2047} & 3.4431 & 3.9446 & 2.2903 & 0.6656 & 0.8427 \\
LLaVA-Next-Video-7B (finetuned) & \textbf{0.1882} & 3.1765 & 3.9433 & \textbf{2.1416} & \textbf{0.6861} & \textbf{0.8663} \\
\midrule
\multicolumn{7}{l}{\textbf{Problem Transformation}} \\
\midrule
PT\_Bayes & 0.9668 {\scriptsize$\pm$ 0.0219} & 21.1994 {\scriptsize$\pm$ 0.0438} & 2.7268 {\scriptsize$\pm$ 0.0477} & 6.9506 {\scriptsize$\pm$ 0.2007} & 0.0144 {\scriptsize$\pm$ 0.0009} & 0.0272 {\scriptsize$\pm$ 0.0015} \\
LDSVR & 0.2584 {\scriptsize$\pm$ 0.0000} & 4.9794 {\scriptsize$\pm$ 0.0000} & \textbf{2.1272} {\scriptsize$\pm$ 0.0000} & 2.8912 {\scriptsize$\pm$ 0.0000} & 0.6146 {\scriptsize$\pm$ 0.0000} & 0.7872 {\scriptsize$\pm$ 0.0000} \\
\midrule
\multicolumn{7}{l}{\textbf{Specialized Algorithms}} \\
\midrule
SA\_BFGS & 0.2306 {\scriptsize$\pm$ 0.0011} & \textbf{0.5976} {\scriptsize$\pm$ 0.0044} & 3.9147 {\scriptsize$\pm$ 0.0012} & \textbf{2.6711} {\scriptsize$\pm$ 0.0146} & 0.6254 {\scriptsize$\pm$ 0.0015} & 0.8089 {\scriptsize$\pm$ 0.0017} \\
LDL\_LRR & 0.3293 {\scriptsize$\pm$ 0.0028} & 2.2569 {\scriptsize$\pm$ 0.0850} & 4.1227 {\scriptsize$\pm$ 0.0007} & 3.5177 {\scriptsize$\pm$ 0.0471} & 0.5242 {\scriptsize$\pm$ 0.0035} & 0.7115 {\scriptsize$\pm$ 0.0127} \\
TLRLDL & 0.3368 {\scriptsize$\pm$ 0.0000} & 7.9606 {\scriptsize$\pm$ 0.0014} & 3.2362 {\scriptsize$\pm$ 0.0001} & 3.8317 {\scriptsize$\pm$ 0.0001} & 0.4264 {\scriptsize$\pm$ 0.0000} & 0.5968 {\scriptsize$\pm$ 0.0000} \\
\midrule
\multicolumn{7}{l}{\textbf{Algorithm Adaptation}} \\
\midrule
CubeMLP & 0.2738 {\scriptsize$\pm$ 0.0002} & 0.6900 {\scriptsize$\pm$ 0.0003} & 3.9669 {\scriptsize$\pm$ 0.0004} & 3.2122 {\scriptsize$\pm$ 0.0003} & 0.5624 {\scriptsize$\pm$ 0.0005} & 0.7513 {\scriptsize$\pm$ 0.0000} \\
CTEN & 0.2432 {\scriptsize$\pm$ 0.0017} & 0.6033 {\scriptsize$\pm$ 0.0044} & 3.9542 {\scriptsize$\pm$ 0.0033} & 2.8277 {\scriptsize$\pm$ 0.0116} & 0.6071 {\scriptsize$\pm$ 0.0021} & 0.7977 {\scriptsize$\pm$ 0.0013} \\
MMIM & 0.2442 {\scriptsize$\pm$ 0.0021} & 0.6076 {\scriptsize$\pm$ 0.0046} & 3.9593 {\scriptsize$\pm$ 0.0009} & 2.8548 {\scriptsize$\pm$ 0.0349} & 0.6019 {\scriptsize$\pm$ 0.0023} & 0.7946 {\scriptsize$\pm$ 0.0027} \\
\bottomrule
\end{tabular}%
}
\label{tab:full_distribution_eval_apx}
\end{table*}

\begin{table*}[ht]
\centering
\caption{Dominant Reaction Evaluation Benchmark Results.}
\resizebox{\textwidth}{!}{%
\begin{tabular}{lccccc}
\toprule
\textbf{Model Name} & TPE $\downarrow$ & MRR $\uparrow$ & F1 Top 1 (weighted) $\uparrow$ & F1 Top 2 (weighted) $\uparrow$ & F1 Top 3 (weighted) $\uparrow$ \\
\midrule
\multicolumn{6}{l}{\textbf{Foundation Video Models}} \\
\midrule
Gemini 2.5 Flash & 0.3026 & 0.4378 & 0.2735 & 0.3332 & 0.3794 \\
LLava-Next-Video-7B & 0.4103 & 0.1992 & 0.0143 & 0.1167 & 0.1374 \\
Qwen2.5-VL & 0.3923 & 0.3088 & 0.1958 & 0.2531 & 0.3037 \\
Qwen2.5-VL (finetuned) & 0.1516 & 0.7548 & \textbf{0.6577} & \textbf{0.7291} & \textbf{0.7725} \\
LLaVA-Next-Video-7B (finetuned) & \textbf{0.1388} & \textbf{0.7833} & \textbf{0.6521} & \textbf{0.7374} & \textbf{0.7672} \\
\midrule
\multicolumn{6}{l}{\textbf{Problem Transformation}} \\
\midrule
PT-Bayes & 0.9661 \scriptsize{± 0.0000} & 0.1535 \scriptsize{± 0.0133} & 0.0001 \scriptsize{± 0.0000} & 0.0031 \scriptsize{± 0.0013} & 0.0741 \scriptsize{± 0.0676} \\
LDSVR & \textbf{0.1599} \scriptsize{± 0.0000} & 0.7054 \scriptsize{± 0.0000} & 0.5034 \scriptsize{± 0.0000} & 0.5865 \scriptsize{± 0.0000} & 0.5696 \scriptsize{± 0.0000} \\
\midrule
\multicolumn{6}{l}{\textbf{Specialized Algorithms}} \\
\midrule
SA-BFGS & 0.1882 \scriptsize{± 0.0021} & 0.7163 \scriptsize{± 0.0038} & 0.5283 \scriptsize{± 0.0053} & 0.6075 \scriptsize{± 0.0014} & 0.6265 \scriptsize{± 0.0026} \\
LDL-LRR & 0.2496 \scriptsize{± 0.0006} & 0.6700 \scriptsize{± 0.0130} & 0.4965 \scriptsize{± 0.0076} & 0.5684 \scriptsize{± 0.0038} & 0.5803 \scriptsize{± 0.0025} \\
TLRLDL & 0.2759 \scriptsize{± 0.0000} & 0.5559 \scriptsize{± 0.0002} & 0.4516 \scriptsize{± 0.0003} & 0.4895 \scriptsize{± 0.0001} & 0.4858 \scriptsize{± 0.0000} \\
\midrule
\multicolumn{6}{l}{\textbf{Algorithm Adaptation}} \\
\midrule
CubeMLP & 0.2509 \scriptsize{± 0.0004} & 0.5996 \scriptsize{± 0.0000} & 0.2376 \scriptsize{± 0.0000} & 0.4399 \scriptsize{± 0.0000} & 0.5587 \scriptsize{± 0.0000} \\
CTEN & 0.2081 \scriptsize{± 0.0022} & 0.6939 \scriptsize{± 0.0022} & 0.4826 \scriptsize{± 0.0062} & 0.5950 \scriptsize{± 0.0022} & 0.5109 \scriptsize{± 0.0013} \\
MMIM & 0.2141 \scriptsize{± 0.0070} & 0.6749 \scriptsize{± 0.0058} & 0.4503 \scriptsize{± 0.0090} & 0.5728 \scriptsize{± 0.0037} & 0.5775 \scriptsize{± 0.0014} \\
\bottomrule
\end{tabular}%
}
\label{tab:dominant_reaction_eval_apx}
\end{table*}

\subsection{Cross-dataset Transfer Learning}
\label{apx:transfer_learning}

We evaluate the transfer learning capability of two vision-language models, Qwen2.5-VL and LLaVA-Next, finetuned on Video2Reaction on a different video dataset with different emotion taxonomy and different demographics of audience, VCE \cite{mazeika2022would}. We investigate several settings: (1) training directly on VCE Dataset, (2) training on \datasetname\ only, and (3) training on \datasetname\ followed by limited supervision from VCE.

Table \ref{tab:vce_transfer} shows that combining \datasetname\ with a small amount of target-domain supervision yields additional improvements. 
With only $1\%$ of VCE training data, Qwen2.5-VL improves to $0.46$ and LLaVA-Next reaches $0.44$. 
These findings indicate that \datasetname\ provides a useful finetuning signal for teaching vision-language models to anticipate audience reactions, and that it can significantly reduce the amount of labeled data required in downstream emotion prediction tasks. 
We leave a more systematic exploration of transfer learning strategies---such as different finetuning objectives, curriculum learning, and multi-stage fine-tuning---to future work.

\begin{table}[t]
\centering
\caption{Transfer learning results on the EmoDiversity (VCE) benchmark using top-3 accuracy. Models fine-tuned on \datasetname\ followed by 1\% of target domain's data demonstrate shows improvements on top-3 accuracy comapred to models fine-tuned on 10\% of target domain's data only.}
\small
\begin{tabular}{lcc}
\hline
\textbf{Training Setup} & \textbf{Qwen2.5-VL} & \textbf{LLaVA-Next} \\
\hline
Random chance & \multicolumn{2}{c}{0.11} \\
Majority emotion & \multicolumn{2}{c}{0.36} \\
VideoMAE (100\% EmoDiversity) & \multicolumn{2}{c}{0.68} \\
\hline
10\% EmoDiversity & 0.35 & 0.37 \\
Video2Reaction only & \textbf{0.41} & 0.31 \\
Video2Reaction + 1\% EmoDiversity & \textbf{0.46} & \textbf{0.44} \\
\hline
\end{tabular}

\label{tab:vce_transfer}
\end{table}

\subsection{Visualization of Predicted Distribution vs. Groundtruth Distribution} 

\begin{figure*}[h]
    \centering
    \includegraphics[width=1\linewidth]{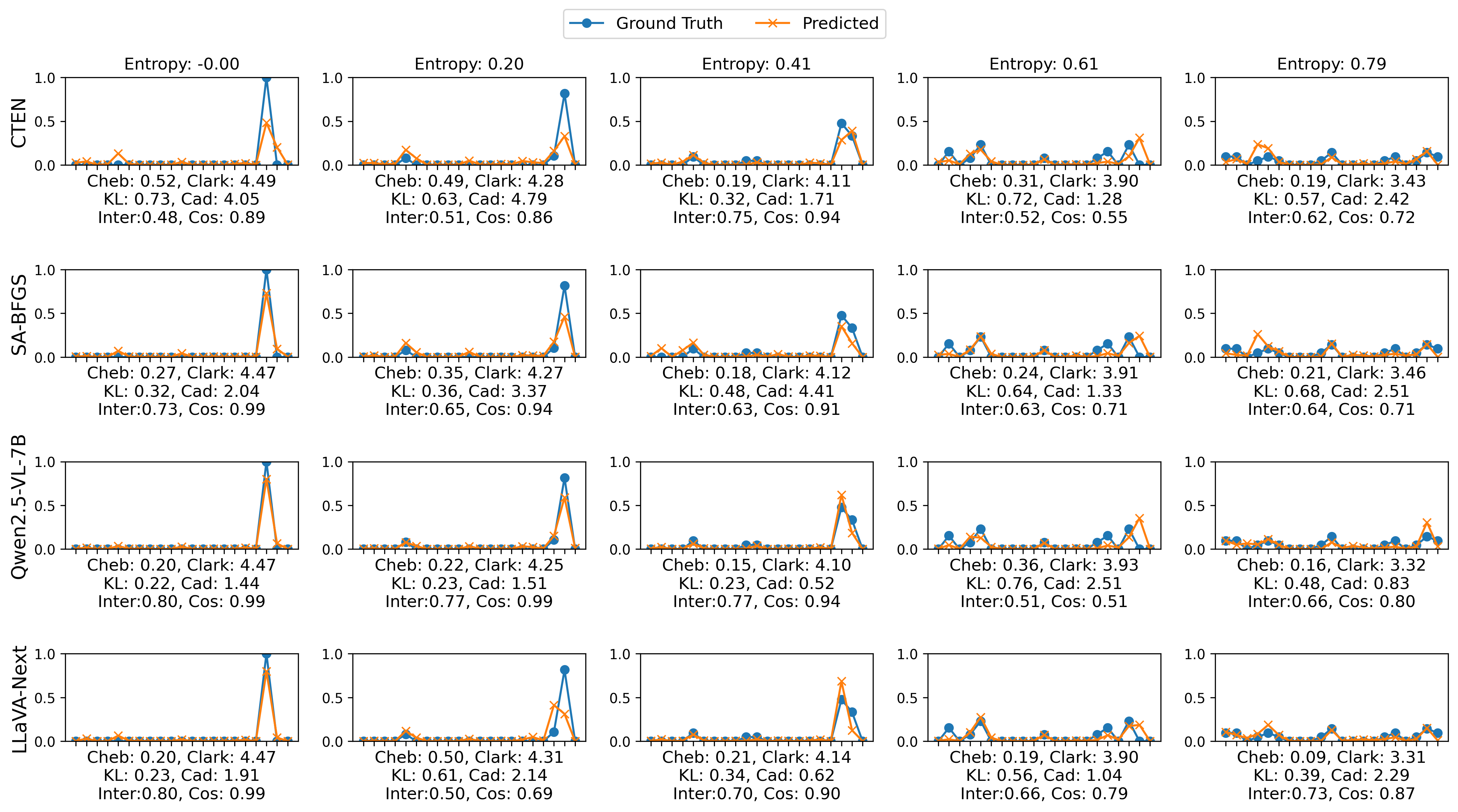}
    \caption{Visual Comparison of Predicted Reaction Distribution of Two Leading Baseline Algorithms. One random sample is selected for each bin of groundtruth distribution entropy. }
    \label{fig:sample_wise_plot}
\end{figure*}

Figure \ref{fig:sample_wise_plot} illustrates how the four leading algorithms in our benchmark (SA-BFGS and CTEN) model groundtruth label distribution across different entropy levels. Lower entropy represents more unimodal distribution and higher entropy represents more uniform distribution. 

\label{apx:viz_samplewise_plot}

\subsection{Long-tail challenge in Dominant Reaction Evaluation} 
Table \ref{tab:reaction_class_f1_scores} shows a breakdown of Top-3 F1 score by each reaction class. 

\label{apx:long_tail}
\begin{table*}[ht]
\centering
\caption{Top-3 F1 Score of finetuned LLaVA-Next by Reaction Class}
\begin{tabular}{lc}
\hline
Reaction Class &  Top-3 F1 Score \\
\hline
annoyance & 0.9995 \\
embarrassment & 0.9985 \\
nervousness & 0.9981 \\
realization & 0.9981 \\
relief & 0.9976 \\
caring & 0.9966 \\
anger & 0.9959 \\
curiosity & 0.9920 \\
grief & 0.9910 \\
joy & 0.9765 \\
excitement & 0.9707 \\
sadness & 0.9665 \\
approval & 0.9660 \\
disgust & 0.9566 \\
fear & 0.9398 \\
disappointment & 0.9321 \\
surprise & 0.9261 \\
confusion & 0.9176 \\
amusement & 0.7136 \\
disapproval & 0.5140 \\
admiration & 0.4918 \\
\hline
\end{tabular}
\label{tab:reaction_class_f1_scores}
\end{table*}

\section{Instruction for Human Review of Automatic Annotations} \label{apx:human_instruction}

\begin{figure*}[ht]
    \centering
    \includegraphics[width=1\linewidth]{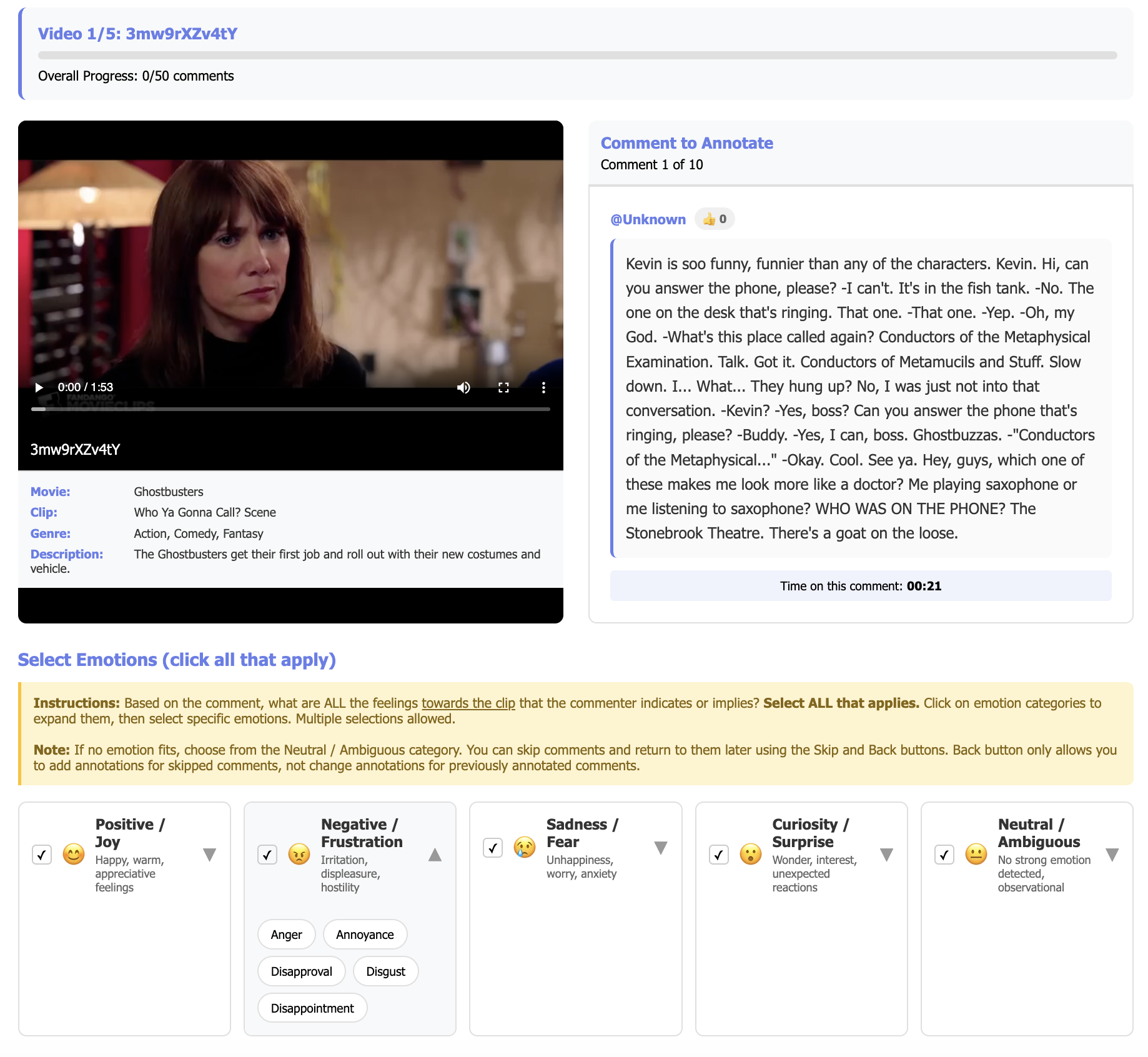}
    \caption{Video-Comment Human Annotaiton Interface}
    \label{fig:human_evaluation_interface.png}
\end{figure*}

\paragraph{Human-LLM Annotation Alignment.} We perform independent human annotation of video-comment pairs with 29 participants. Human annotators perform their annotation via an annotation tool as shown in \cref{fig:human_evaluation_interface.png}. Each participant has to complete a tutorial of labeling 7 comments to make sure that we can get high-quality annotations. 

\paragraph{Dual Blind Human Verification}. To assess the quality of automated reaction annotation, we randomly sample 100 movie clips with balanced representation across all movie genres. From each clip, 10 comments are randomly selected, yielding a total of 1,000 comments for human evaluation. Due to the subjective nature of fine-grained audience reactions, each comment is independently reviewed by two annotators. In cases of disagreement, a third annotator is consulted, and the final label is determined by majority vote. In total, five annotators participated in this quality review process. To account for confirmation bias, we present reviewers with either LLM-labeled or randomly-labeled answers. Only $0.2\%$ of randomly-labeled answers are judged to be correct. Instruction for the annotation is described below.

\begin{tcolorbox}[colback=gray!10!white, colframe=gray!30!black,  title=Instruction for Human Reviewers, boxrule=0.5pt, arc=3pt, before skip=10pt, after skip=10pt, sharp corners]
The LLM agent was tasked with annotating the fine-grained reaction of the YouTube commenter towards the movie clip.

For each comment, choose one of the following ratings:

\begin{itemize}
    \item \textbf{Correct}: You think there is a high chance that the LLM response is correct. Sometimes the model may stretch to infer the audience's intent, but if you believe the guess is reasonable, rate it as Correct.
    
    \item \textbf{Incorrect}: You think there is a low chance that the LLM response is correct. If the model's guess seems too far-fetched or if a different reaction label clearly fits better, rate it as Incorrect.
    
    \item \textbf{Not Sure}: You cannot understand the comment well enough to make a judgment.
\end{itemize}
\end{tcolorbox}

\end{document}